%% file: main.tex
\documentclass[lettersize, journal]{IEEEtran}
\usepackage{amsmath, amsfonts, amssymb}
\usepackage{algorithmic}
\usepackage{algorithm}
\usepackage{array}
\usepackage[caption=false, font=normalsize, labelfont=sf, textfont=sf]{subfig}
\usepackage{textcomp}
\usepackage{stfloats}
\usepackage{url}
\usepackage{verbatim}
\usepackage{graphicx}
\usepackage{cite}
\usepackage{amsthm}
\usepackage{tikz}
\usepackage{booktabs}
\usepackage{multirow}   % For \multirow (merging rows)
\usepackage{array} %
\usepackage{makecell}
\usepackage{hyperref}
\usepackage{xcolor}
\definecolor{R1}{named}{black}

\newcommand{\rev}[1]{\textcolor{R1}{#1}}
\begin{document}
\title{Impedance Control of Ship-Borne Manipulators via Optimization-based Task-Space Inverse Dynamics}

% \author{\IEEEauthorblockN{Anonymous Submission}
% }

\author{Lingxiao Meng, Bi-Ke Zhu, Xuheng Gao, Zhe Zhang, Jiankun Yang, Jiankun Wang, 
Haibo Lu, Max Q.-H. Meng
}

\markboth{Journal of \LaTeX\ Class Files,~Vol.~14, No.~8, August~2021}%
{Shell \MakeLowercase{\textit{et al.}}: A Sample Article Using IEEEtran.cls for IEEE Journals}


\maketitle

\input{intro.tex}
\input{tsid.tex}

\input{eskf.tex}
\input{exp.tex}

\input{conclusion.tex}
\input{appendix.tex}

\bibliographystyle{IEEEtran}
\bibliography{references}

\vfill
\end{document}

%% file: intro.tex
\begin{abstract}
Ship-borne manipulators operating in maritime environments are subject to stochastic wave-induced base motions that introduce kinematic disturbances and dynamic coupling, degrading trajectory tracking accuracy and complicating safe, contact-rich manipulation.
This paper proposes a torque-level optimization-based control framework that integrates high-precision trajectory tracking with task-space impedance for ship-borne manipulators. The controller is formulated using task-space inverse dynamics (TSID) and solved via quadratic programming to explicitly compensate for the dynamic coupling introduced by base motion. To enable accurate feedforward compensation, an error-state Kalman filter (ESKF) is developed to estimate the base state by fusing inertial measurements with end-effector pose feedback.
The framework is validated in simulation and real-world experiments using a 7-DOF manipulator mounted on a 6-DOF Stewart platform. The proposed method reduces real-world end-effector position tracking error by over 25.7\% compared with the best baseline. Furthermore, the controller enables dynamic peg-in-hole insertion with 1~mm clearance under base motion, increasing the success rate while reducing average contact forces by \rev{45}\%, demonstrating precise and compliant manipulation in contact-rich environments.
\end{abstract}

\def\abstractname{Note to Practitioners}
\begin{abstract}
Ship-borne manipulators are essential for automating offshore tasks such as refueling unmanned vessels, equipment maintenance, and assisting unmanned aerial vehicle landings. A major practical challenge in these applications is that wind and ocean waves constantly rock the ship. This motion significantly reduces manipulator precision, making physical contact tasks, such as inserting connectors or grasping objects, difficult and potentially unsafe. Existing methods struggle to maintain both accuracy and safe interaction in such environments.
This paper presents a control framework designed for manipulators mounted on ships. The method computes motor torques that explicitly compensate for base motion while enforcing a compliant, spring-like behavior. As a result, the manipulator can track targets accurately while remaining tolerant to contact, reducing the risk of damaging the robot or the surrounding environment. The framework also includes a real-time state estimator that fuses inertial sensor data with robot feedback to track ship motion.
A practical limitation is the reliance on accurate robot dynamic models; performance may degrade if payload properties or robot dynamic parameters are uncertain. Future work will investigate adaptive techniques for handling unknown payloads and evaluate the approach in field experiments.
\end{abstract}


\section{Introduction}

Marine robots play an important role in ocean exploration and automated offshore operations \cite{skaugset2025,from2011motion}.
Ship-borne manipulators extend the operational capability of marine platforms by enabling a variety of tasks such as unmanned aerial vehicle (UAV) landing assistance \cite{xu2024tro, su2025coordinated}, aquaculture maintenance \cite{brandt2023towards}, unmanned surface vehicle (USV) refueling \cite{scott2015autonomous}, and floating object recovery \cite{li2024modeling}.
In such environments, the ship undergoes stochastic 6-DOF motions induced by waves. 
These disturbances significantly degrade manipulation accuracy and stability, making reliable task execution in maritime environments particularly challenging.

Specifically, base motion introduces three fundamental challenges: 1) the continuously varying base pose induces relative pose variation between the manipulator and the target, making precise positioning difficult; 2) base motion introduces dynamic coupling into the manipulator system, which can significantly degrade trajectory tracking performance if not properly compensated; and 3) during contact-rich operations, base-induced kinematic misalignments increase the risk of unintended contacts, potentially generating large impact forces that destabilize manipulation or damage the system.
These factors make it challenging to simultaneously achieve accurate tracking and compliant physical interaction in ship-borne manipulation.

To address the first challenge of relative pose variation, several studies have developed motion compensation strategies at the kinematic level. \cite{tordal2017} and \cite{brandt2024automated} employ inverse kinematics (IK) to compute joint configurations that keep the end-effector stationary relative to a target. To better account for ship motion, model predictive control (MPC) has been employed to predict future base motions and optimize the end-effector velocity for inertial-frame tracking \cite{peec}. \cite{li2025optimized} and \cite{li2025motion} employ visual servoing to compute desired Cartesian velocity commands based on pose errors, while neural networks predict ship motion for active compensation.
In addition, \cite{evolver} employs a data-driven disturbance observer to estimate and compensate for velocity disturbances induced by base motion. 
\rev{
These methods can effectively reduce geometric tracking errors caused by the moving base because they explicitly regulate the relative pose or velocity between the manipulator and the target while accounting for ship motion.
}

\rev{
However, they do not explicitly model and compensate for the second challenge posed by base-induced dynamic coupling.
In practice, the velocity and acceleration of the moving base introduce significant inertial forces into the manipulator dynamics, rendering standard fixed-base dynamic models inadequate for precise control \cite{from2009modeling}.  
Because these methods usually rely on low-level joint-velocity controllers designed for fixed-base operation, the base-induced torque disturbances must be rejected indirectly by the underlying velocity servo, which may limit achievable tracking accuracy.
}

\rev{
Beyond purely kinematic compensation, some studies attempt to mitigate the dynamic influence of base motion at the velocity-control level.  
For example, \cite{xuoe} represents the base-induced torque disturbance as an equivalent joint-velocity disturbance.
A disturbance observer is therefore used to estimate the base-induced joint velocity disturbance, which is then added to the joint velocity command.  
This approach improves tracking performance in UAV landing assistance, but both disturbance estimation and compensation are performed at the velocity level and therefore depend on the underlying velocity servo.
}

\rev{
Torque-level dynamic control provides a unified framework in which the geometric effects of base motion are compensated at the task-space level, while the associated dynamic coupling is addressed directly in the manipulator dynamics.
Several studies model the inertial forces induced by base motion as external disturbances in the manipulator dynamics.
}
For example, \cite{er2023composite} employs sliding-mode control to compensate for such disturbances in the manipulator dynamics, while \cite{su2025appointed} uses \rev{an} observer-based \rev{approach} to estimate and compensate for disturbance torques introduced by base motion. These methods demonstrate improved tracking performance in simulation.

Beyond achieving high tracking accuracy, ship-borne manipulators must also maintain stable physical interaction during contact-rich operations. For tasks such as rescue manipulation, hydraulic structure inspection, or USV charging, compliant behavior is essential to mitigate rigid impacts and prevent damage to both the manipulator and the environment \cite{wang2025robust, park2017compliance}. Unfortunately, the aforementioned approaches are generally unsuitable for such scenarios.
Velocity-based control frameworks typically require high gains to achieve accurate velocity tracking, which limits compliance. Existing dynamic disturbance-rejection controllers also have difficulty distinguishing between inertial forces and external contact forces, leading to unstable behavior during contact. To introduce compliance, \cite{vasiljevic2025robust} implements admittance control that adjusts the Cartesian velocity based on force/torque sensor feedback for ship-to-ship object grasping. Nevertheless, admittance control relies heavily on the performance of the underlying velocity controller and may exhibit reduced stability under highly dynamic contact conditions compared with impedance control \cite{5509861}.

To address these limitations, we propose an optimization-based inverse dynamics control framework for ship-borne manipulators. The control objective is formulated as a task-space inverse dynamics (TSID) problem \cite{del2016robustness} and solved via quadratic programming (QP). 
\rev{
The proposed framework accounts for base motion at both the task-space kinematic and joint-space dynamic levels. The base-induced velocity and acceleration terms are included in the task-space acceleration model, while the corresponding inertial torques are explicitly compensated through feedforward torque commands.
}
This improves trajectory-tracking accuracy while enforcing a virtual mass-spring-damper behavior at the end-effector for safe interaction during contact tasks.
Accurate feedforward compensation requires a reliable real-time estimate of the base motion. However, accurately measuring the full base state in ship-borne manipulation systems is often difficult due to limited sensing \cite{cao2022nonlinear}. 
In contrast, the end-effector pose is typically more readily available, as many practical manipulation systems rely on eye-in-hand perception to provide visual feedback during task execution. Motivated by this observation, we develop an error-state Kalman filter (ESKF) \cite{sola2017quaternion} that fuses measurements from a base-mounted IMU with end-effector pose feedback to estimate the base state in real time.

The main contributions of this paper are summarized as follows:
\begin{itemize}
\item We propose a TSID-based control framework formulated as a QP that explicitly compensates for base-induced dynamic coupling. The torque-level controller achieves accurate trajectory tracking under base motion while enforcing task-space impedance behavior for safe interaction with the environment.
\item We develop a real-time state estimator based on an ESKF. By fusing measurements from a base-mounted IMU with end-effector pose feedback, the estimator provides an accurate base state required for dynamic compensation.
\item We validate the proposed framework through experiments on a 6-DOF Stewart platform that emulates ship motion. Comparative results demonstrate that the proposed method improves trajectory tracking performance. Furthermore, in a challenging dynamic peg-in-hole insertion task, the proposed method achieves a higher success rate while significantly reducing contact forces compared with baseline methods.
\end{itemize}

The remainder of this paper is organized as follows. Section II presents the TSID control framework. Section III introduces the ESKF method for base state estimation. Section IV validates the proposed approach through simulation and real-world experiments. Section V discusses system limitations. Section VI concludes the paper.

%% file: tsid.tex
\section{Inverse Dynamics Control}

\subsection{Ship-borne Manipulator Dynamics}

\begin{figure}[t]
    \centering
    \includegraphics[width=3.2in]{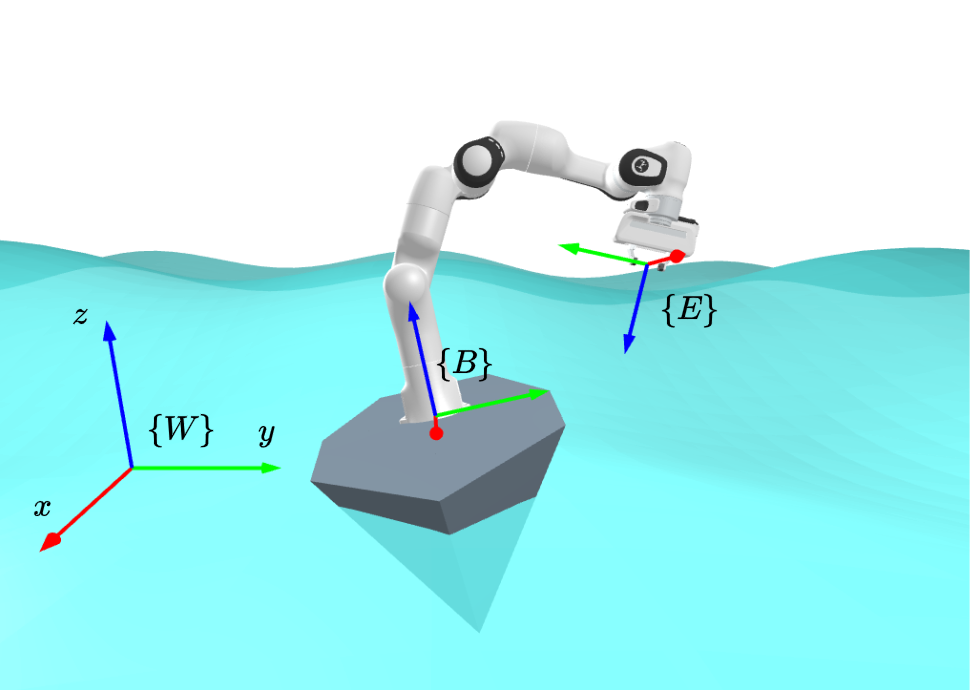}
    \caption{Coordinate frames for a ship-borne manipulator.}
    \label{fig:coordinate_frames}
\end{figure}

Unlike fixed-base manipulators, ship-borne manipulators are subject to stochastic base disturbances. The system dynamics \rev{couple} the manipulator's joint states with the ship's 6-DOF motion. Let $\{W\}$ denote the inertial world frame, $\{B\}$ the base frame attached to the ship, and $\{E\}$ the end-effector frame (see Fig. \ref{fig:coordinate_frames}). The rigid-body dynamics of an $n$-DOF ship-borne manipulator are modeled as:
\begin{equation} 
    \boldsymbol{M}(\boldsymbol{q})\ddot{\boldsymbol{q}} + \boldsymbol{C}(\boldsymbol{q},\dot{\boldsymbol{q}})\dot{\boldsymbol{q}} + \boldsymbol{g}(\boldsymbol{q}, \boldsymbol{R}_{B}^W) + \boldsymbol{\tau}_{\text{base}} = \boldsymbol{\tau} + \boldsymbol{\tau}_{\text{ext}} + \boldsymbol{\tau}_{d}, \label{eq:dynamics} 
\end{equation}
where $\boldsymbol{q}, \dot{\boldsymbol{q}}, \ddot{\boldsymbol{q}} \in \mathbb{R}^{n}$ are the joint position, velocity, and acceleration vectors.
$\boldsymbol{M}(\boldsymbol{q}) \in \mathbb{R}^{n\times n}$ is the inertia matrix, $\boldsymbol{C}(\boldsymbol{q},\dot{\boldsymbol{q}}) \in \mathbb{R}^{n\times n}$ is the Coriolis and centrifugal matrix, and $\boldsymbol{g}(\boldsymbol{q}, \boldsymbol{R}_{B}^W) \in \mathbb{R}^{n}$ is the gravity vector, which depends on the joint position $\boldsymbol{q}$ and the base orientation $\boldsymbol{R}_{B}^W \in SO(3)$ relative to $\{W\}$.
The joint control torque is $\boldsymbol{\tau} \in \mathbb{R}^{n}$, and unmodeled disturbances (e.g., friction and model uncertainties) are lumped into $\boldsymbol{\tau}_{d} \in \mathbb{R}^{n}$. The \rev{joint} torque induced by \rev{the} external contact \rev{wrench} $\boldsymbol{F}_{\text{ext}} \in \mathbb{R}^6$ applied at $\{E\}$ is $\boldsymbol{\tau}_{\text{ext}} = \boldsymbol{J}^T\boldsymbol{F}_{\text{ext}}$, where $\boldsymbol{J} \in \mathbb{R}^{6\times n}$ is the manipulator Jacobian.

The term $\boldsymbol{\tau}_{\text{base}}\in \mathbb{R}^{n}$ represents the disturbance torques induced by ship motion:
\begin{equation}
\boldsymbol{\tau}_{\text{base}} = \boldsymbol{M}_{B}(\boldsymbol{q}) \dot{\mathcal{V}}_{B}^{B}+ \boldsymbol{C}_{B}(\boldsymbol{q}, \dot{\boldsymbol{q}}, \mathcal{V}_{B}^{B}) \mathcal{V}_{B}^{B}, \label{eq:base_coupling}
\end{equation}
where $\mathcal{V}_{B}^B, \dot{\mathcal{V}}_{B}^B \in \mathbb{R}^6$ are the spatial velocity and acceleration of the base, expressed in $\{B\}$. The matrices $\boldsymbol{M}_{B}(\boldsymbol{q}) \in \mathbb{R}^{n \times 6}$ and $\boldsymbol{C}_{B}(\boldsymbol{q}, \dot{\boldsymbol{q}}, \mathcal{V}_{B}^{B}) \in \mathbb{R}^{n \times 6}$ represent the dynamic coupling between the base and the manipulator joints. $\boldsymbol{\tau}_{\text{base}}$ can be computed efficiently via the recursive Newton-Euler algorithm (RNEA) \cite{featherstone2008rigid}.
\rev{
When the base is fixed, $\mathcal{V}_{B}^B=0$ and $\dot{\mathcal{V}}_{B}^B=0$, and thus $\boldsymbol{\tau}_{\text{base}}$ vanishes, and the model reduces to the standard fixed-base dynamics. 
For a ship-borne manipulator, the nonzero base velocity and acceleration generate additional inertial torques at the manipulator joints, represented by $\boldsymbol{\tau}_{\text{base}}$. 
If not explicitly compensated, the effect of this term must be attenuated through feedback, which may degrade tracking performance. 
Therefore, $\boldsymbol{\tau}_{\mathrm{base}}$ is explicitly included in the inverse dynamics model and compensated through feedforward torque commands.
}

\begin{figure}[tb]
    \centering
    \includegraphics[width=\linewidth]{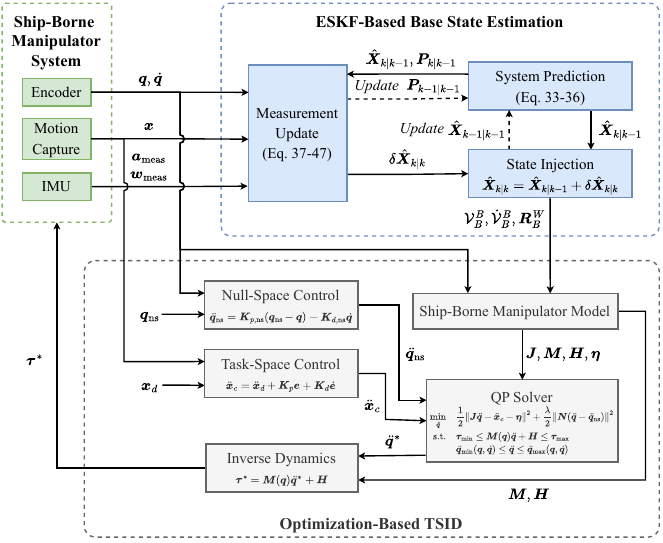}
    \caption{Overview of the proposed control framework.}
    \label{fig:control}
\end{figure}

\subsection{Optimization-Based TSID}

To unify trajectory tracking and dynamic compliance, a TSID framework is formulated as a QP problem. The controller \rev{computes} optimal joint accelerations and torques to satisfy task-space objectives.

Let $\boldsymbol{x} \in SE(3)$ and $\dot{\boldsymbol{x}} \in \mathbb{R}^6$ denote the pose and spatial velocity of the end-effector in $\{W\}$. Based on differential kinematics, the end-effector velocity is given by:
\begin{equation}
    \label{eq:fk}\dot{\boldsymbol{x}}= \boldsymbol{J}\dot{\boldsymbol{q}}
    + \boldsymbol{J}_{B}\mathcal{V}_{B}^{B},
\end{equation}
where $\boldsymbol{J}_{B} \in \mathbb{R}^{6\times 6}$ is the base Jacobian matrix mapping the base velocity in $\{B\}$ to the end-effector velocity in $\{W\}$. Differentiating \eqref{eq:fk} yields the end-effector spatial acceleration $\ddot{\boldsymbol{x}} \in \mathbb{R}^6$:

\begin{equation}
    \ddot{\boldsymbol{x}}= \underbrace{\boldsymbol{J}\ddot{\boldsymbol{q}} +
    \dot{\boldsymbol{J}}\dot{\boldsymbol{q}}}_{\text{Manipulator
    Contribution}}+ \underbrace{\boldsymbol{J}_{B}\dot{\mathcal{V}}_{B}^{B}
    + \dot{\boldsymbol{J}}_{B}\mathcal{V}_{B}^{B}}_{\text{Base Motion Contribution}}.
\end{equation}

To track a reference trajectory $\boldsymbol{x}_d(t) \in SE(3)$, a virtual command acceleration $\ddot{\boldsymbol{x}}_c \in \mathbb{R}^6$ is synthesized using a proportional-derivative (PD) control law defined on the $SE(3)$ manifold. The task-space pose error $\boldsymbol{e} \in \mathbb{R}^6$ is defined via the logarithmic map:
\begin{equation}
    \boldsymbol{e} = \log(\boldsymbol{x}^{-1}\boldsymbol{x}_d),
\end{equation}
where $\log(\cdot): SE(3) \to \mathfrak{se}(3)$ maps the relative transformation to a 6D error vector.
The command acceleration is then given by:
\begin{equation}
    \ddot{\boldsymbol{x}}_c = \ddot{\boldsymbol{x}}_d + \boldsymbol{K}_p \boldsymbol{e} + \boldsymbol{K}_d \dot{\boldsymbol{e}},
\end{equation}
where $\boldsymbol{K}_p, \boldsymbol{K}_d \in \mathbb{R}^{6\times 6}$ are positive-definite stiffness and damping gain matrices and $\dot{\boldsymbol{e}} = \dot{\boldsymbol{x}}_d - \dot{\boldsymbol{x}}$ is the task-space velocity error.

To achieve $\ddot{\boldsymbol{x}} = \ddot{\boldsymbol{x}}_c$, the joint accelerations must compensate for task-space drift induced by nonlinear kinematics and base motion. The required constraint on joint accelerations is:
\begin{equation}
    \boldsymbol{J}\ddot{\boldsymbol{q}} = \ddot{\boldsymbol{x}}_c - \boldsymbol{\eta},
\end{equation}
where $\boldsymbol{\eta} = \dot{\boldsymbol{J}}\dot{\boldsymbol{q}} + \boldsymbol{J}_{B}{}\dot{\mathcal{V}}_{B}^{B} + \dot{\boldsymbol{J}}_{B}\mathcal{V}_{B}^{B}$ represents the task-space drift acceleration.

For a kinematically redundant manipulator ($n > 6$), the additional degrees of freedom can be used to optimize secondary objectives, such as singularity avoidance. 
\rev{When these objectives are projected} into the Jacobian null space, these secondary tasks do not affect the primary end-effector trajectory tracking.
In practice, these objectives are achieved by keeping the joints near a safe, nominal posture. Accordingly, a secondary PD control law drives the manipulator toward a nominal joint position $\boldsymbol{q}_{\text{ns}} \in \mathbb{R}^n$, generating a reference null-space acceleration $\ddot{\boldsymbol{q}}_{\text{ns}} \in \mathbb{R}^n$:
\begin{equation}
    \ddot{\boldsymbol{q}}_{\text{ns}} = \boldsymbol{K}_{p, \text{ns}} (\boldsymbol{q}_{\text{ns}} - \boldsymbol{q}) - \boldsymbol{K}_{d, \text{ns}} \dot{\boldsymbol{q}},
\end{equation}
where $\boldsymbol{K}_{p, \text{ns}}, \boldsymbol{K}_{d, \text{ns}} \in \mathbb{R}^{n \times n}$ are positive-definite diagonal gain matrices.

The optimal joint acceleration $\ddot{\boldsymbol{q}}$ is \rev{obtained} by minimizing the tracking errors of both primary and secondary tasks. To ensure safe, physically feasible execution, the optimization is constrained by system dynamics and actuator torque limits $[\boldsymbol{\tau}_{\min}, \boldsymbol{\tau}_{\max}]$. Furthermore, joint position and velocity limits, $[\boldsymbol{q}_{\min}, \boldsymbol{q}_{\max}]$ and $[\dot{\boldsymbol{q}}_{\min}, \dot{\boldsymbol{q}}_{\max}]$, are mapped to feasible acceleration bounds $[\ddot{\boldsymbol{q}}_{\min}, \ddot{\boldsymbol{q}}_{\max}]$ using a state-prediction scheme parameterized by the control sampling period $\Delta t$:
\begin{equation}
    \begin{cases}
        \ddot{\boldsymbol{q}}_{\max}= \min \left(\dfrac{\dot{\boldsymbol{q}}_{\max} - \dot{\boldsymbol{q}}}{\Delta t}, \ \dfrac{2(\boldsymbol{q}_{\max} -\boldsymbol{q} - \dot{\boldsymbol{q}}\Delta t)}{\Delta t^2}\right)  \\
        \ddot{\boldsymbol{q}}_{\min}= \max \left( \dfrac{\dot{\boldsymbol{q}}_{\min} -\dot{\boldsymbol{q}}}{\Delta t}, \ \dfrac{2(\boldsymbol{q}_{\min} - \boldsymbol{q} - \dot{\boldsymbol{q}}\Delta t)}{\Delta t^2}\right).
    \end{cases}
\end{equation}

The final TSID problem is formulated as a QP:
\begin{equation}\label{eq:qp_objective}
    \begin{aligned}\min_{\ddot{\boldsymbol{q}}}\quad &\frac{1}{2}\left\| \boldsymbol{J}\ddot{\boldsymbol{q}} - \ddot{\boldsymbol{x}}_c
    + \boldsymbol{\eta}\right\|^{2}+ \frac{\lambda}{2}
    \left\| \boldsymbol{N}(\ddot{\boldsymbol{q}}- \ddot{\boldsymbol{q}}_{\text{ns}}
    ) \right\|^{2} \\
    \text{s.t.} \quad & \boldsymbol{\tau}_{\min} \le \boldsymbol{M}(\boldsymbol{q})\ddot{\boldsymbol{q}} + \boldsymbol{H} \le \boldsymbol{\tau}_{\max} \\
    & \ddot{\boldsymbol{q}}_{\min}(\boldsymbol{q}, \dot{\boldsymbol{q}}) \le \ddot{\boldsymbol{q}} \le \ddot{\boldsymbol{q}}_{\max}(\boldsymbol{q}, \dot{\boldsymbol{q}})
    \end{aligned}
\end{equation}
where $\boldsymbol{H} = \boldsymbol{C}(\boldsymbol{q},\dot{\boldsymbol{q}})\dot{\boldsymbol{q}} + \boldsymbol{g}(\boldsymbol{q}, \boldsymbol{R}_{B}^W) + \boldsymbol{\tau}_{\text{base}}$ encompasses the nonlinear dynamics, and $\lambda$ is a weighting factor.
The first term in the objective function enforces precise tracking of the end-effector trajectory, while the second term minimizes the \rev{null-space component of the deviation from $\ddot{\boldsymbol{q}}_{\text{ns}}$}.
The kinematic null-space projector is defined as:
\begin{equation}
    \boldsymbol{N} = \boldsymbol{I} - \boldsymbol{J}^\dagger \boldsymbol{J},
\end{equation}
where $\boldsymbol{J}^\dagger \in \mathbb{R}^{n \times 6}$ is the Moore-Penrose pseudoinverse of $\boldsymbol{J}$, and $\boldsymbol{I} \in \mathbb{R}^{n \times n}$ is the identity matrix. The matrix $\boldsymbol{N} \in \mathbb{R}^{n \times n}$ projects joint-space vectors onto the null space of $\boldsymbol{J}$.

At each control cycle, the QP solver outputs the optimal joint accelerations $\ddot{\boldsymbol{q}}^{*}$. The corresponding command torques $\boldsymbol{\tau}^*$ are then computed via the inverse dynamics model:
\begin{equation}
    \boldsymbol{\tau}^*= \boldsymbol{M}(\boldsymbol{q})\ddot{\boldsymbol{q}}^{*} + \boldsymbol{H}.
\end{equation}

\subsection{Analysis of Task-Space Impedance Properties}

\label{subsec:impedance_analysis}

{\color{R1}

To analyze the task-space impedance properties, the QP in \eqref{eq:qp_objective} is written as: 
\begin{equation}
    \begin{aligned}
    \min_{\ddot{\boldsymbol{q}}}\quad
    & \frac{1}{2}\ddot{\boldsymbol{q}}^T
    \boldsymbol{Q}\ddot{\boldsymbol{q}}
    -
    \boldsymbol{h}^T\ddot{\boldsymbol{q}} \\
    \mathrm{s.t.}\quad
    & \boldsymbol{A}\ddot{\boldsymbol{q}} \leq \boldsymbol{b},
    \end{aligned}
    \label{eq:qp_standard}
\end{equation}
where constant terms independent of $\ddot{\boldsymbol{q}}$ have been omitted, and
\begin{equation}
    \boldsymbol{Q}
    =
    \boldsymbol{J}^T\boldsymbol{J}
    +
    \lambda \boldsymbol{N}^T\boldsymbol{N},
\end{equation}
\begin{equation}
    \boldsymbol{h}
    =
    \boldsymbol{J}^T
    \left(
    \ddot{\boldsymbol{x}}_c-\boldsymbol{\eta}
    \right)
    +
    \lambda
    \boldsymbol{N}^T\boldsymbol{N}
    \ddot{\boldsymbol{q}}_{\mathrm{ns}}.
\end{equation}
The constraint matrices are 
\begin{equation}
    \boldsymbol{A}
    =
    \begin{bmatrix}
    \boldsymbol{M} \\
    -\boldsymbol{M} \\
    \boldsymbol{I} \\
    -\boldsymbol{I}
    \end{bmatrix},
    \qquad
    \boldsymbol{b}
    =
    \begin{bmatrix}
    \boldsymbol{\tau}_{\max}-\boldsymbol{H} \\
    \boldsymbol{H}-\boldsymbol{\tau}_{\min} \\
    \ddot{\boldsymbol{q}}_{\max} \\
    -\ddot{\boldsymbol{q}}_{\min}
    \end{bmatrix}.
    \label{eq:qp_constraint_matrices}
\end{equation}

When no inequality constraint is active, the solution reduces to:
\begin{equation}
    \ddot{\boldsymbol{q}}^{*}
    =
    \ddot{\boldsymbol{q}}_0
    =
    \boldsymbol{J}^{\dagger}
    \left(
    \ddot{\boldsymbol{x}}_c-\boldsymbol{\eta}
    \right)
    +
    \boldsymbol{N}\ddot{\boldsymbol{q}}_{\mathrm{ns}} .
    \label{eq:qdd_unconstrained}
\end{equation}

When inequality constraints are active, the constrained solution generally deviates from $\ddot{\boldsymbol{q}}_0$. Let $\mathcal{I}$ denote the active constraint set, and let $\boldsymbol{A}_{\mathcal{I}}$ and $\boldsymbol{b}_{\mathcal{I}}$ denote the corresponding rows of $\boldsymbol{A}$ and $\boldsymbol{b}$, respectively. For a given active set, the KKT conditions are
\begin{equation}
    \begin{bmatrix}
    \boldsymbol{Q} & \boldsymbol{A}_{\mathcal{I}}^T \\
    \boldsymbol{A}_{\mathcal{I}} & \boldsymbol{0}
    \end{bmatrix}
    \begin{bmatrix}
    \ddot{\boldsymbol{q}}^{*} \\
    \boldsymbol{\mu}_{\mathcal{I}}
    \end{bmatrix}
    =
    \begin{bmatrix}
    \boldsymbol{h} \\
    \boldsymbol{b}_{\mathcal{I}}
    \end{bmatrix},
    \qquad
    \boldsymbol{\mu}_{\mathcal{I}}\geq \boldsymbol{0},
    \label{eq:kkt_active}
\end{equation}
where $\boldsymbol{\mu}_{\mathcal{I}}$ is the vector of Lagrange multipliers associated with the active constraints. From the stationarity condition, the constrained solution can be written as:
\begin{equation}
    \ddot{\boldsymbol{q}}^{*}
    =
    \ddot{\boldsymbol{q}}_0
    +
    \Delta\ddot{\boldsymbol{q}}_c,
    \qquad
    \Delta\ddot{\boldsymbol{q}}_c
    =
    -
    \boldsymbol{Q}^{-1}
    \boldsymbol{A}_{\mathcal{I}}^T
    \boldsymbol{\mu}_{\mathcal{I}}.
    \label{eq:qdd_constrained_decomposition}
\end{equation}

Here, $\Delta\ddot{\boldsymbol{q}}_c$ is the joint-acceleration correction induced by the active constraints. If no inequality constraint is active, then $\boldsymbol{\mu}_{\mathcal{I}}=\boldsymbol{0}$ and $\Delta\ddot{\boldsymbol{q}}_c=\boldsymbol{0}$.

Although $\lambda$ does not affect the unconstrained solution $\ddot{\boldsymbol{q}}_0$, it influences the constrained correction through $\boldsymbol{Q}$ and the Lagrange multipliers. 
When constraints are active, a smaller $\lambda$ allows larger changes in the null-space acceleration, which may help satisfy the constraints with a smaller deviation in the primary task. A larger $\lambda$ keeps the null-space acceleration closer to $\ddot{\boldsymbol{q}}_{\mathrm{ns}}$, but may require a larger task-space correction $\boldsymbol{J}\Delta\ddot{\boldsymbol{q}}_c$.

Applying the control torque $\boldsymbol{\tau}^* = \boldsymbol{M}(\boldsymbol{q})\ddot{\boldsymbol{q}}^{*} + \boldsymbol{H}$ to the actual system dynamics \eqref{eq:dynamics} in the presence of external wrench $\boldsymbol{F}_{\text{ext}}$ and disturbances $\boldsymbol{\tau}_d$ yields:
\begin{equation}
\boldsymbol{M}\ddot{\boldsymbol{q}} + \boldsymbol{H} = \boldsymbol{M}\ddot{\boldsymbol{q}}^{*} + {\boldsymbol{H}} + \boldsymbol{\tau}_{\text{ext}} + \boldsymbol{\tau}_{d}.
\end{equation}

The joint-space closed-loop dynamics simplify to:
\begin{equation}
\boldsymbol{M}(\ddot{\boldsymbol{q}} - \ddot{\boldsymbol{q}}^{*}) = \boldsymbol{J}^{T}\boldsymbol{F}_{\text{ext}} + \boldsymbol{\tau}_{d}.
\end{equation}

Solving for $\ddot{\boldsymbol{q}}$ and pre-multiplying by the Jacobian $\boldsymbol{J}$ gives:
\begin{equation}
    \boldsymbol{J}\ddot{\boldsymbol{q}} = \boldsymbol{J}\ddot{\boldsymbol{q}}^{*} + \boldsymbol{J}\boldsymbol{M}^{-1}(\boldsymbol{J}^{T}\boldsymbol{F}_{\text{ext}} + \boldsymbol{\tau}_{d}).
\end{equation}

Substituting $\ddot{\boldsymbol{q}}^{*}$ from \eqref{eq:qdd_constrained_decomposition}:

\begin{equation}
\boldsymbol{J}\ddot{\boldsymbol{q}} = \boldsymbol{J}\left( \ddot{\boldsymbol{q}}_0 + \Delta\ddot{\boldsymbol{q}}_c \right) + \boldsymbol{J}\boldsymbol{M}^{-1}(\boldsymbol{J}^{T}\boldsymbol{F}_{\text{ext}} + \boldsymbol{\tau}_{d}).
\end{equation}

Using $\boldsymbol{J}\boldsymbol{N}=\boldsymbol{0}$ and
$\boldsymbol{J}\boldsymbol{J}^{\dagger}=\boldsymbol{I}$ away from kinematic singularities, the contribution of the secondary task $\ddot{\boldsymbol{q}}_{\text{ns}}$ vanishes from the task-space dynamics and the above equation simplifies to:
\begin{equation}
\boldsymbol{J}\ddot{\boldsymbol{q}} = \ddot{\boldsymbol{x}}_c - \boldsymbol{\eta} + \boldsymbol{J}\Delta\ddot{\boldsymbol{q}}_c + \boldsymbol{J}\boldsymbol{M}^{-1}(\boldsymbol{J}^{T}\boldsymbol{F}_{\text{ext}} + \boldsymbol{\tau}_{d}).
\end{equation}

Using the differential kinematic relation $\ddot{\boldsymbol{x}} = \boldsymbol{J}\ddot{\boldsymbol{q}} + \boldsymbol{\eta}$, we obtain:
\begin{equation}
\ddot{\boldsymbol{x}}  = \ddot{\boldsymbol{x}}_{c} + \boldsymbol{J}\Delta\ddot{\boldsymbol{q}}_c + \boldsymbol{J}\boldsymbol{M}^{-1}\boldsymbol{J}^{T}\boldsymbol{F}_{\text{ext}} + \boldsymbol{J}\boldsymbol{M}^{-1}\boldsymbol{\tau}_{d}.
\end{equation}

Defining the operational-space inertia matrix $\boldsymbol{\Lambda}(\boldsymbol{q}) = (\boldsymbol{J}\boldsymbol{M}^{-1}\boldsymbol{J}^{T})^{-1}$ and pre-multiplying both sides by $\boldsymbol{\Lambda}$ yields:
\begin{equation}
    \boldsymbol{\Lambda}
    \left(
    \ddot{\boldsymbol{x}}
    -
    \ddot{\boldsymbol{x}}_c
    \right)
    =
    \boldsymbol{F}_{\mathrm{ext}}
    +
    \boldsymbol{\delta}_d
    +
    \boldsymbol{\delta}_c ,
    \label{eq:constrained_impedance_model}
\end{equation}
where $\boldsymbol{\delta}_d=\boldsymbol{\Lambda}\boldsymbol{J}\boldsymbol{M}^{-1}\boldsymbol{\tau}_d$ is the equivalent task-space wrench induced by unmodeled joint-space disturbances, and $\boldsymbol{\delta}_c=\boldsymbol{\Lambda}\boldsymbol{J}\Delta\ddot{\boldsymbol{q}}_c$ is the equivalent task-space wrench induced by active inequality constraints.

After substituting the virtual command acceleration $\ddot{\boldsymbol{x}}_{c} = \ddot{\boldsymbol{x}}_{d} + \boldsymbol{K}_{p}\boldsymbol{e} + \boldsymbol{K}_{d}\dot{\boldsymbol{e}}$ and applying the spatial error relations $\dot{\boldsymbol{e}} = \dot{\boldsymbol{x}}_d - \dot{\boldsymbol{x}}$ and $\ddot{\boldsymbol{e}} = \ddot{\boldsymbol{x}}_d - \ddot{\boldsymbol{x}}$, the closed-loop error dynamics become:
\begin{equation}
    \boldsymbol{\Lambda}\ddot{\boldsymbol{e}}
    +
    \boldsymbol{\Lambda}\boldsymbol{K}_d\dot{\boldsymbol{e}}
    +
    \boldsymbol{\Lambda}\boldsymbol{K}_p\boldsymbol{e}
    =
    -
    \boldsymbol{F}_{\mathrm{ext}}
    -
    \boldsymbol{\delta}_d
    -
    \boldsymbol{\delta}_c .
    \label{eq:constrained_impedance_error}
\end{equation}

Equation \eqref{eq:constrained_impedance_error} shows that the end-effector follows a task-space virtual mass--spring--damper error model. The external wrench $\boldsymbol{F}_{\mathrm{ext}}$ represents physical interaction, while $\boldsymbol{\delta}_d$ represents residual model uncertainty and $\boldsymbol{\delta}_c$ represents the effect of active inequality constraints. Hence, the end-effector responds compliantly to $\boldsymbol{F}_{\mathrm{ext}}$ when both additional terms are small.

When no inequality constraint is active, $\boldsymbol{\delta}_c=\boldsymbol{0}$, and the deviation from the desired impedance dynamics is determined mainly by $\boldsymbol{\delta}_d$. With the identified dynamic parameters, $\boldsymbol{\delta}_d$ is expected to remain small under nominal operating conditions. When inequality constraints become active, a nonzero $\boldsymbol{\delta}_c$ is introduced and may increase near joint-state or torque limits. Consequently, the actual task-space behavior may deviate from the desired impedance dynamics, resulting in degraded tracking and compliance performance.

As shown in Appendix~\ref{app:uub_proof}, if the disturbances are bounded, the QP remains feasible, and the manipulator operates away from kinematic singularities, $\boldsymbol{\delta}_d$ and $\boldsymbol{\delta}_c$ are bounded. Therefore, the tracking error is uniformly ultimately bounded \cite{khalil2002nonlinear}. This result, however, guarantees boundedness rather than small tracking error. If $\boldsymbol{\delta}_c$ becomes large, the ultimate bound may also become large and may no longer be meaningful for practical operation. This limitation indicates that the current formulation is mainly applicable when the inequality constraints are inactive or only mildly active.
}

%% file: eskf.tex
\section{ESKF-Based Base State Estimation}
\label{sec:eskf_estimation}

Precise control of a ship-borne manipulator requires accurate real-time estimation of the base motion. 
In practical scenarios, visual feedback is often obtained using an eye-in-hand configuration. Compared with a camera mounted on the ship, this configuration reduces the relative motion between the camera and the target, mitigates motion blur, and helps maintain the target within the field of view, thereby providing more reliable feedback for manipulator control. 
Motivated by this sensor setup, we propose an ESKF-based estimation method that fuses inertial measurements from a base-mounted IMU with end-effector pose feedback to estimate the base state in real time. Since the primary focus of this work is the control framework rather than the visual perception pipeline, the end-effector pose is obtained from a motion capture system, which provides accurate and stable pose measurements for state estimation and control evaluation.

\subsection{State Definition}
The \rev{true} state vector $\boldsymbol{X}$ is defined as:
\begin{equation}
    \boldsymbol{X} =
    \begin{bmatrix}
        \boldsymbol{p}^{{W}}_{{B}},
        \boldsymbol{\xi}^{{W}}_{{B}},
        \boldsymbol{v}^{{B}}_{{B}},
        \boldsymbol{\omega}^{{B}}_{{B}},
        \boldsymbol{a}^{{B}}_{{B}},
        \boldsymbol{\alpha}^{{B}}_{{B}},
        \boldsymbol{b}_{a},
        \boldsymbol{b}_{\omega},
        \boldsymbol{p}^{{W}}_{E},
        \boldsymbol{\xi}^{{W}}_{E}
    \end{bmatrix}^{T},
\end{equation}
where $\boldsymbol{p}^{{W}}_{{B}} \in \mathbb{R}^3$ and $\boldsymbol{\xi}^{{W}}_{{B}} \in \mathbb{S}^3$ denote the base position and unit quaternion; $\boldsymbol{v}^{{B}}_{{B}}, \boldsymbol{\omega}^{{B}}_{{B}}, \boldsymbol{a}_{{B}}^{{B}}, \boldsymbol{\alpha}_{{B}}^{{B}} \in \mathbb{R}^3$ represent the base linear velocity, angular velocity, linear acceleration, and angular acceleration, all expressed in $\{B\}$; $\boldsymbol{b}_{a}, \boldsymbol{b}_{\omega} \in \mathbb{R}^3$ are the accelerometer and gyroscope biases; and $\boldsymbol{p}^{{W}}_{E} \in \mathbb{R}^3, \boldsymbol{\xi}^{{W}}_{E} \in \mathbb{S}^3$ are the end-effector position and orientation in $\{W\}$.

Following the standard ESKF formulation, the corresponding error-state vector
$\delta \boldsymbol{X} \in \mathbb{R}^{30}$ is defined as:
\begin{equation}
\begin{aligned}
    \delta \boldsymbol{X} =\big[
        &\delta \boldsymbol{p}^{{W}}_{{B}},
        \delta \boldsymbol{\theta}^{{B}}_{{B}},
        \delta \boldsymbol{v}^{{B}}_{{B}},
        \delta \boldsymbol{\omega}^{{B}}_{{B}},
        \delta \boldsymbol{a}^{{B}}_{{B}}, \\
        &\delta \boldsymbol{\alpha}^{{B}}_{{B}},
        \delta \boldsymbol{b}_{a},
        \delta \boldsymbol{b}_{\omega},
        \delta \boldsymbol{p}^{{W}}_{E},
        \delta \boldsymbol{\theta}^{{E}}_{E}
    \big]^{T},
\end{aligned}
\end{equation}
where $\delta \boldsymbol{\theta} \in \mathbb{R}^3$ denotes the minimal representation of the orientation error in the tangent space of $SO(3)$. For \rev{the true} quaternion $\boldsymbol{\xi}$ and \rev{its nominal estimate} $\hat{\boldsymbol{\xi}}$, the orientation error is given by:
\begin{equation}
    \boldsymbol{\xi} =  \hat{\boldsymbol{\xi}} \otimes \delta \boldsymbol{\xi}, \quad
    \delta \boldsymbol{\xi} = \exp(\delta \boldsymbol{\theta}) \approx
    \begin{bmatrix}
        1 \\
        \tfrac{1}{2}\delta \boldsymbol{\theta}
    \end{bmatrix},
\end{equation}
where $\otimes$ denotes quaternion multiplication, and $\exp(\cdot)$ maps the small-angle rotation vector to a unit quaternion.

\subsection{System Prediction Model}

The continuous-time state differential equations, corresponding to the system without noise, can be formulated as:

\begin{equation}
    \begin{cases}
    \dot{\boldsymbol{p}}^{{W}}_{{B}} = \boldsymbol{R}^{{W}}_{{B}} \boldsymbol{v}^{{B}}_{{B}}, \\
    \dot{\boldsymbol{\xi}}^{{W}}_{{B}} =\frac{1}{2} \boldsymbol{\xi}^{{W}}_{{B}} \otimes \begin{bmatrix} 0 \\ \boldsymbol{\omega}^{{B}}_{{B}} \end{bmatrix}, \\
    \dot{\boldsymbol{v}}^{{B}}_{{B}} = \boldsymbol{a}^{{B}}_{{B}}
    - \lfloor \boldsymbol{\omega}^{{B}}_{{B}} \rfloor_{\times} \boldsymbol{v}^{{B}}_{{B}}, \\
    \dot{\boldsymbol{\omega}}^{{B}}_{{B}} = \boldsymbol{\alpha}^{{B}}_{{B}}, \\
    \dot{\boldsymbol{a}}^{{B}}_{{B}} = \boldsymbol{0}, \\
    \dot{\boldsymbol{\alpha}}^{{B}}_{{B}} = \boldsymbol{0}, \\
    \dot{\boldsymbol{b}}_a = \boldsymbol{0},\\
    \dot{\boldsymbol{b}}_\omega = \boldsymbol{0},\\
    \dot{\boldsymbol{p}}^{{W}}_{E} =
    \boldsymbol{R}^{{W}}_{{B}}
    \left(
        \boldsymbol{v}^{{B}}_{{B}}
        + \lfloor \boldsymbol{\omega}^{{B}}_{{B}} \rfloor_{\times} \boldsymbol{p}^{{B}}_{E}
        + \boldsymbol{v}^{{B}}_{\text{arm}}
    \right),\\
    \dot{\boldsymbol{\xi}}^{{W}}_{E} = \frac{1}{2} \boldsymbol{\xi}^{{W}}_{E} \otimes \begin{bmatrix} 0 \\ \boldsymbol{R}^{E}_{{B}} (\boldsymbol{\omega}^{{B}}_{{B}}
    + \boldsymbol{\omega}^{{B}}_{\text{arm}}) \end{bmatrix}
    ,
\end{cases}
\end{equation}
where $\boldsymbol{p}^{{B}}_{E}$,  $\boldsymbol{v}^{{B}}_{\text{arm}}$, and $\boldsymbol{\omega}^{{B}}_{\text{arm}}$ are the position, linear velocity, and angular velocity of the end-effector expressed in $\{B\}$. The velocities $\boldsymbol{v}^{{B}}_{\text{arm}}$ and $\boldsymbol{\omega}^{{B}}_{\text{arm}}$ correspond to the end-effector motion induced by the manipulator joints and are obtained from the forward kinematics \rev{(FK)}.
\rev{The linear and angular accelerations are included as explicit states under a locally constant-acceleration model. Their temporal variations are represented by process noise. This design allows the estimator to directly provide the velocity and acceleration required by the controller without additional processing.}

The operator $\lfloor \cdot \rfloor_{\times}$ denotes the skew-symmetric matrix operator, which represents the cross product in matrix form. For a vector $\boldsymbol{\omega} = [\omega_x, \omega_y, \omega_z]^T$, the skew-symmetric matrix is defined as:
\begin{equation}
    \lfloor\boldsymbol{\omega}\rfloor_{\times} =
    \begin{bmatrix}
        0 & -\omega_{z} & \omega_{y} \\
        \omega_{z} & 0 & -\omega_{x} \\
        -\omega_{y} & \omega_{x} & 0
    \end{bmatrix}.
\end{equation}

The prediction step of the ESKF numerically integrates these \rev{state} equations over a sampling interval $\Delta t$ to propagate the nominal state:
\begin{equation}
    \begin{cases}
    \boldsymbol{p}^{{W}}_{{B}} = \boldsymbol{p}^{{W}}_{{B}} + \boldsymbol{R}^{{W}}_{{B}} \boldsymbol{v}^{{B}}_{{B}} \Delta t, \\
    \boldsymbol{\xi}^{{W}}_{{B}} = \boldsymbol{\xi}^{{W}}_{{B}} \otimes \exp(\boldsymbol{\omega}^{{B}}_{{B}} \Delta t), \\
    \boldsymbol{v}^{{B}}_{{B}} = \boldsymbol{v}^{{B}}_{{B}} + \left( \boldsymbol{a}^{{B}}_{{B}}
    - \lfloor \boldsymbol{\omega}^{{B}}_{{B}} \rfloor_{\times} \boldsymbol{v}^{{B}}_{{B}} \right) \Delta t, \\
    \boldsymbol{\omega}^{{B}}_{{B}} = \boldsymbol{\omega}^{{B}}_{{B}} + \boldsymbol{\alpha}^{{B}}_{{B}} \Delta t, \\
    \boldsymbol{a}^{{B}}_{{B}} = \boldsymbol{a}^{{B}}_{{B}}, \\
    \boldsymbol{\alpha}^{{B}}_{{B}} = \boldsymbol{\alpha}^{{B}}_{{B}}, \\
    \boldsymbol{b}_a = \boldsymbol{b}_a,\\
    \boldsymbol{b}_\omega = \boldsymbol{b}_\omega,\\
    \boldsymbol{p}^{{W}}_{E} = \boldsymbol{p}^{{W}}_{E} +
    \boldsymbol{R}^{{W}}_{{B}}
    \left(
        \boldsymbol{v}^{{B}}_{{B}}
        + \lfloor \boldsymbol{\omega}^{{B}}_{{B}} \rfloor_{\times} \boldsymbol{p}^{{B}}_{E}
        + \boldsymbol{v}^{{B}}_{\text{arm}}
    \right) \Delta t,
    \\
    \boldsymbol{\xi}^{{W}}_{E} = \boldsymbol{\xi}^{{W}}_{E} \otimes
    \exp
    \left(\boldsymbol{R}^{E}_{{B}} (\boldsymbol{\omega}^{{B}}_{{B}}
    + \boldsymbol{\omega}^{{B}}_{\text{arm}})\Delta t\right).
    \end{cases}
\end{equation}

The above equations propagate the nominal state \rev{from
$\hat{\boldsymbol{X}}_{k-1|k-1}$ to the predicted estimate
$\hat{\boldsymbol{X}}_{k|k-1}$.}

To propagate the error covariance, the linearized noise-free error-state dynamics are derived as:
\begin{equation}
\begin{cases}
    \delta \dot{\boldsymbol{p}}^{W}_{B} = \boldsymbol{R}^{W}_{B} \delta \boldsymbol{v}^{B}_{B} - \boldsymbol{R}^{W}_{B} \lfloor \boldsymbol{v}^{B}_{B} \rfloor_{\times} \delta \boldsymbol{\theta}^{B}_{B}, \\
    \delta \dot{\boldsymbol{\theta}}^{B}_{B} = \delta \boldsymbol{\omega}^{B}_{B} - \lfloor \boldsymbol{\omega}^{B}_{B} \rfloor_{\times} \delta \boldsymbol{\theta}^{B}_{B}, \\
    \delta \dot{\boldsymbol{v}}^{B}_{B} = \delta \boldsymbol{a}^{B}_{B} - \lfloor \boldsymbol{\omega}^{B}_{B} \rfloor_{\times} \delta \boldsymbol{v}^{B}_{B} + \lfloor \boldsymbol{v}^{B}_{B} \rfloor_{\times} \delta \boldsymbol{\omega}^{B}_{B}, \\
    \delta \dot{\boldsymbol{\omega}}^{B}_{B} = \delta \boldsymbol{\alpha}^{B}_{B}, \\
    \delta \dot{\boldsymbol{a}}^{B}_{B} = \boldsymbol{0}, \\
    \delta \dot{\boldsymbol{\alpha}}^{B}_{B} = \boldsymbol{0}, \\
    \delta \dot{\boldsymbol{b}}_a = \boldsymbol{0}, \\
    \delta \dot{\boldsymbol{b}}_\omega = \boldsymbol{0}, \\
    \delta \dot{\boldsymbol{p}}^{W}_{E} = \boldsymbol{R}^{W}_{B} \big( \delta \boldsymbol{v}^{B}_{B} - \lfloor \boldsymbol{p}^{B}_{E} \rfloor_{\times} \delta \boldsymbol{\omega}^{B}_{B} \\ 
    \quad\quad\quad\,\,\, - \lfloor \boldsymbol{v}^{B}_{B} + \lfloor \boldsymbol{\omega}^{B}_{B} \rfloor_{\times} \boldsymbol{p}^{B}_{E} + \boldsymbol{v}^{B}_{\text{arm}} \rfloor_{\times} \delta \boldsymbol{\theta}^{B}_{B} \big), \\
    \delta \dot{\boldsymbol{\theta}}^{E}_{E} = \boldsymbol{R}^{E}_{B} \delta \boldsymbol{\omega}^{B}_{B} - \lfloor \boldsymbol{R}^{E}_{B}(\boldsymbol{\omega}^{B}_{B} + \boldsymbol{\omega}^{B}_{\text{arm}}) \rfloor_{\times} \delta \boldsymbol{\theta}^{E}_{E}.
\end{cases}
\end{equation}

This system of linear differential equations can be expressed compactly as:
\begin{equation}
    \delta \dot{\boldsymbol{X}} = \boldsymbol{F}_{c}\delta \boldsymbol{X} + \boldsymbol{n},
\end{equation}
where $\boldsymbol{F}_{c} \in \mathbb{R}^{30 \times 30}$
is the continuous-time error-state Jacobian matrix, and $\boldsymbol{n}\in \mathbb{R}^{30}$ is the process noise vector, modeled as zero-mean Gaussian with covariance $\boldsymbol{Q}_{c}$.

The discrete-time state transition matrix is approximated as $\boldsymbol{\Phi} \approx \boldsymbol{I} + \boldsymbol{F}_{c}\Delta t$. Error covariance propagation is executed via standard Kalman filter prediction equations:
\begin{equation}
    \boldsymbol{P}_{k|k-1} = \boldsymbol{\Phi}_{k} \boldsymbol{P}_{k-1|k-1} \boldsymbol{\Phi}_{k}^{T} + \boldsymbol{Q}_{d},
\end{equation}
where $\boldsymbol{Q}_{d} = \boldsymbol{Q}_{c}\Delta t$ is the discrete-time process noise covariance matrix.

\subsection{Measurement Model}

Due to the disparate sampling rates of the sensors, the filter operates in a multi-rate framework. The IMU data and the end-effector pose feedback are treated as asynchronous updates, applied whenever a valid measurement arrives.

The general nonlinear measurement model is defined as:
\begin{equation}
\boldsymbol{z}_k = \boldsymbol{h}(\boldsymbol{X}_k) + \boldsymbol{n}_k,
\label{eq:measurement_model}
\end{equation}
where $\boldsymbol{z}_k$ is the measurement vector at time step $k$, $\boldsymbol{h}(\cdot)$ is the nonlinear observation function, and $\boldsymbol{n}_k \sim \mathcal{N}(\boldsymbol{0}, \boldsymbol{R}_k)$ is zero-mean Gaussian measurement noise with covariance $\boldsymbol{R}_k$.

Linearizing the measurement model around the predicted state estimate $\hat{\boldsymbol{X}}_{k|k-1}$ yields the error-state observation model:
\begin{equation}
    \delta\boldsymbol{z}_k = \boldsymbol{H}_k \delta\boldsymbol{X}_k + \boldsymbol{n}_k,
    \label{eq:linearized_measurement}
\end{equation}
where $\delta\boldsymbol{z}_k = \boldsymbol{z}_k - \boldsymbol{h}(\hat{\boldsymbol{X}}_{k|k-1})$ is the measurement residual, $\delta\boldsymbol{X}_k$ is the error state, and $\boldsymbol{H}_k = \left. \frac{\partial \boldsymbol{h}}{\partial \boldsymbol{X}} \right|_{\hat{\boldsymbol{X}}_{k|k-1}}$ is the measurement Jacobian.

The ESKF update utilizes the standard Kalman equations:
\begin{align}
    \boldsymbol{K}_k &= \boldsymbol{P}_{k|k-1} \boldsymbol{H}_k^\top \left( \boldsymbol{H}_k \boldsymbol{P}_{k|k-1} \boldsymbol{H}_k^\top + \boldsymbol{R}_k \right)^{-1}, \label{eq:kalman_gain} \\
    \delta \hat{\boldsymbol{X}}_{k|k} &= \boldsymbol{K}_k \delta \boldsymbol{z}_k, \label{eq:state_update} \\
    \boldsymbol{P}_{k|k} &= \left( \boldsymbol{I} - \boldsymbol{K}_k \boldsymbol{H}_k \right) \boldsymbol{P}_{k|k-1}, \label{eq:covariance_update}
\end{align}
where $\boldsymbol{K}_k$ is the Kalman gain, $\delta\hat{\boldsymbol{X}}_{k|k}$ is the estimated error state correction, and $\boldsymbol{P}_{k|k}$ is the updated error covariance matrix. 

Following the update, the nominal state estimate is corrected by injecting the estimated error state $\delta \hat{\boldsymbol{X}}_{k|k}$:
\begin{equation}
    \hat{\boldsymbol{X}}_{k|k} = \hat{\boldsymbol{X}}_{k|k-1} + \delta \hat{\boldsymbol{X}}_{k|k}.
\end{equation} 

Note that for the orientation components, both the measurement residual computation and the final error-state injection are performed on the Lie group manifold rather than using standard Euclidean arithmetic.

\subsubsection{IMU Measurement}
In contrast to standard methods that utilize IMU data purely for prediction, the accelerometer and gyroscope measurements are treated as observations.
This structure facilitates data smoothing and direct estimation of base accelerations.
The IMU measurement vector $\boldsymbol{z}_\text{imu} \in \mathbb{R}^{6}$ is defined as:
\begin{equation}
\boldsymbol{z}_{\text{imu}} = \begin{bmatrix} \boldsymbol{a}_{\text{meas}} \\ \boldsymbol{\omega}_{\text{meas}} \end{bmatrix} = \begin{bmatrix} \boldsymbol{a}^{B}_{B} + {\boldsymbol{R}^{B}_{W}} \boldsymbol{g}^{W} + \boldsymbol{b}_{a} + \boldsymbol{n}_{a} \\ \boldsymbol{\omega}^{B}_{B} + \boldsymbol{b}_{\omega} + \boldsymbol{n}_{\omega} \end{bmatrix},
\end{equation}
where $\boldsymbol{g}^{{W}}$ is the gravity vector in the inertial frame. The corresponding Jacobian $\boldsymbol{H}_\text{imu} \in \mathbb{R}^{6 \times 30}$ maps the error state to the residual:
\begin{equation}
    \boldsymbol{H}_\text{imu} = 
    \begin{bmatrix}
        \boldsymbol{0} & \lfloor\boldsymbol{R}^{B}_{W}\boldsymbol{g}^{{W}}\rfloor_{\times} & \boldsymbol{0} & \boldsymbol{0} & \boldsymbol{I} & \boldsymbol{0} & \boldsymbol{I} & \boldsymbol{0} & \boldsymbol{0} & \boldsymbol{0} \\
        \boldsymbol{0} & \boldsymbol{0} & \boldsymbol{0} & \boldsymbol{I} & \boldsymbol{0} & \boldsymbol{0} & \boldsymbol{0} & \boldsymbol{I} & \boldsymbol{0} & \boldsymbol{0}
    \end{bmatrix}.
\end{equation}

\subsubsection{End-effector Pose Measurement}
To improve the observability of the base state, an augmented measurement vector $\boldsymbol{z}_{\text{pose}}$ is constructed by combining the directly observed end-effector pose and a synthesized base pose obtained through \rev{FK}:
\begin{equation}
    \boldsymbol{z}_{\text{pose}} = \begin{bmatrix} \boldsymbol{p}_{E, \text{meas}}^{W} \\ \boldsymbol{\xi}_{E, \text{meas}}^{W} \\ \boldsymbol{p}_{B, \rev{fk}}^{W} \\ \boldsymbol{\xi}_{B, \rev{fk}}^{W} \end{bmatrix},
\end{equation}
where $\boldsymbol{p}_{E,\text{meas}}^{W}$ and $\boldsymbol{\xi}_{E,\text{meas}}^{W}$ denote the measured position and orientation of the end-effector in the world frame $\{W\}$. 
The terms $\boldsymbol{p}_{B,\rev{fk}}^{W}$ and $\boldsymbol{\xi}_{B,\rev{fk}}^{W}$ denote the synthesized base pose \rev{inferred from the measured end-effector pose and the relative end-effector pose computed from the manipulator forward kinematics.}

The Jacobian $\boldsymbol{H}_{\text{pose}} \in \mathbb{R}^{12 \times 30}$ maps the error states to the measurement residuals.
\begin{equation}
    \boldsymbol{H}_{\text{pose}} = 
    \begin{bmatrix}
        \boldsymbol{0}_{6 \times 6} & \boldsymbol{0}_{6 \times 18} & \boldsymbol{I}_{6 \times 6} \\
        \boldsymbol{I}_{6 \times 6} & \boldsymbol{0}_{6 \times 18} & \boldsymbol{0}_{6 \times 6}
    \end{bmatrix}.
\end{equation}

The first block row corresponds to the direct end-effector measurement, and the second block row corresponds to the \rev{FK}-derived base measurement. The reliability of these two measurement sources differs significantly because the derived base pose accumulates errors along the kinematic chain. To account for this, the measurement noise covariance matrix is modeled as a block-diagonal matrix:
\begin{equation}
\boldsymbol{R}_{\text{pose}} =
\begin{bmatrix}
\boldsymbol{R}_E & \boldsymbol{0} \\
\boldsymbol{0} & \boldsymbol{R}_{\rev{fk}}
\end{bmatrix},
\end{equation}
where $\boldsymbol{R}_E$ represents the covariance of the direct end-effector measurement, and $\boldsymbol{R}_{\rev{fk}}$ corresponds to the uncertainty of the \rev{FK}-derived base pose. In practice, we set $\boldsymbol{R}_{\rev{fk}} \gg \boldsymbol{R}_E$ to reflect the higher uncertainty of the pseudo-measurement.

%% file: exp.tex
\section{Experiments}

In this section, \rev{the proposed framework is evaluated through simulation and real-world experiments. 
We first compare its trajectory-tracking performance with baseline controllers and analyze the effects of constraint activation and model uncertainty. 
We then evaluate its real-time computational performance and base-state estimation accuracy.
Finally, dynamic peg-in-hole experiments are conducted to assess contact performance and insertion reliability under base motion.}

\subsection{Simulation Trajectory Tracking Experiments}

\subsubsection{Setup and Implementation}
A simulation platform is developed using Gazebo and ROS Noetic. The 7-DOF robot model and controller interface are implemented based on the \emph{franka\_gazebo} package.
To emulate ship-induced base motion, a controllable 6-DOF virtual joint is introduced between the robot base and the world frame. IMU measurements are generated using the Gazebo \emph{imu\_sensor} plugin. Ground-truth poses of the robot base and the end-effector are obtained directly from the Gazebo physics engine for quantitative evaluation.

The proposed control framework is implemented in C++ and ROS, running at 1 kHz on a desktop computer with an Intel \rev{i9}-13900K processor. Rigid-body dynamics are computed using the Pinocchio library \cite{pinocchio}, and the QP formulations are \rev{solved} using the TSID library \cite{tsid}. The manipulator's dynamic parameters correspond to the identified URDF model from \cite{panda}.
\begin{figure}[htbp]
    \centering
    \includegraphics[width=\linewidth]{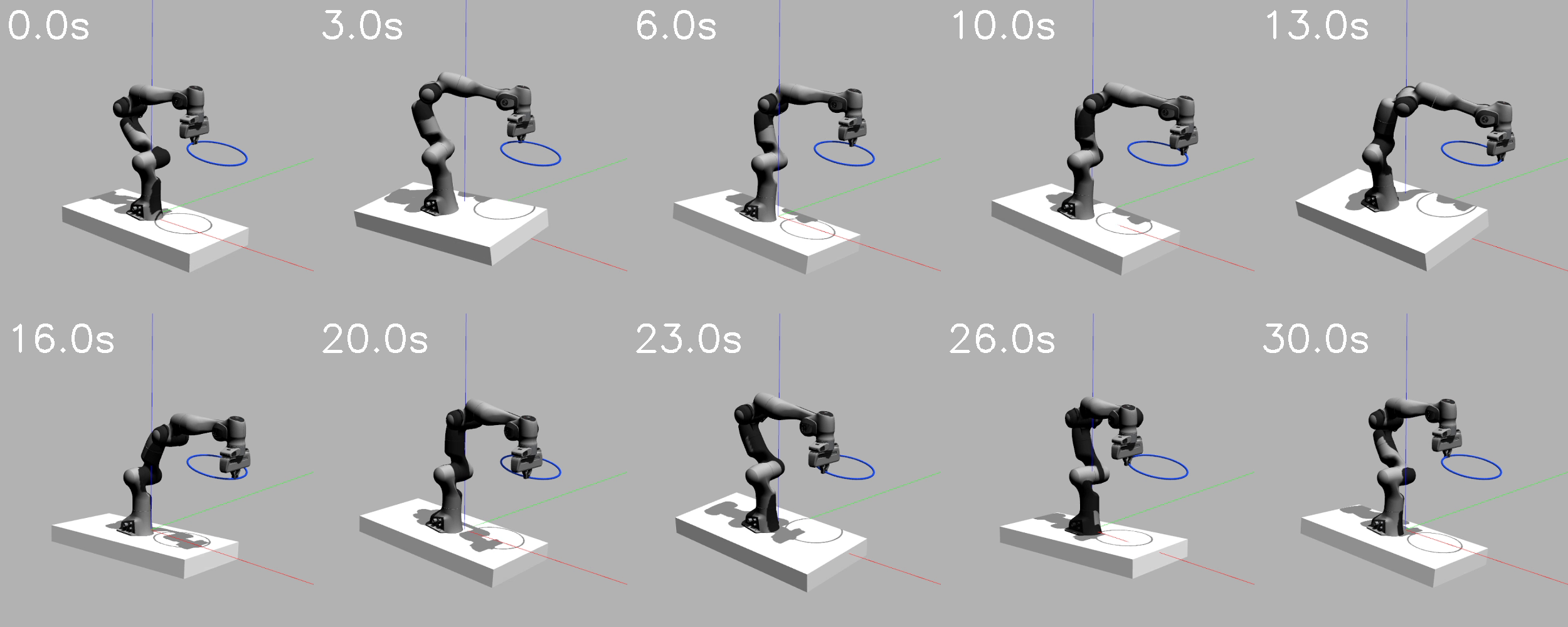}
    \caption{Snapshot of the circular trajectory tracking experiment in simulation.}
    \label{fig:sim_video}
\end{figure}
\begin{figure}[htbp]
    \centering
    \includegraphics[width=\linewidth]{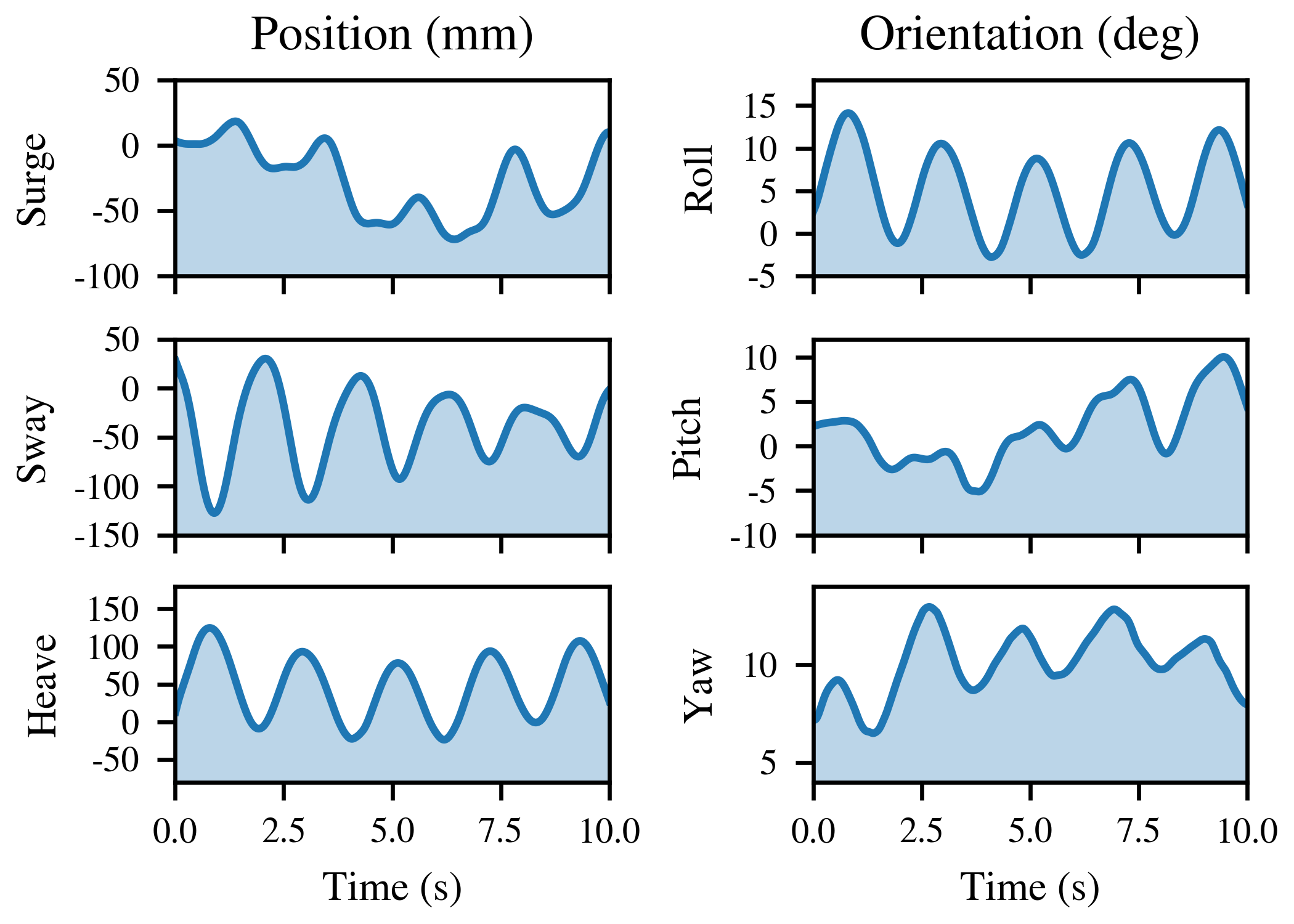}
    \caption{Base motion for simulation experiments.}
    \label{fig:sim_base_motion}
\end{figure}
\subsubsection{Task Description and Base Motion}
The manipulator is tasked with tracking predefined end-effector trajectories while the base undergoes dynamic motion.
We evaluate three tasks: (1) fixed-point stabilization (10 s duration), (2) circular tracking ($r = 0.15$ m, 30 s duration), and (3) figure-8 tracking defined by $x = 0.15\sin(t)$ and $y = 0.225\sin(t)\cos(t)$ (30 s duration).

\begin{figure*}[tbp]
    \centering
    \includegraphics[width=\linewidth]{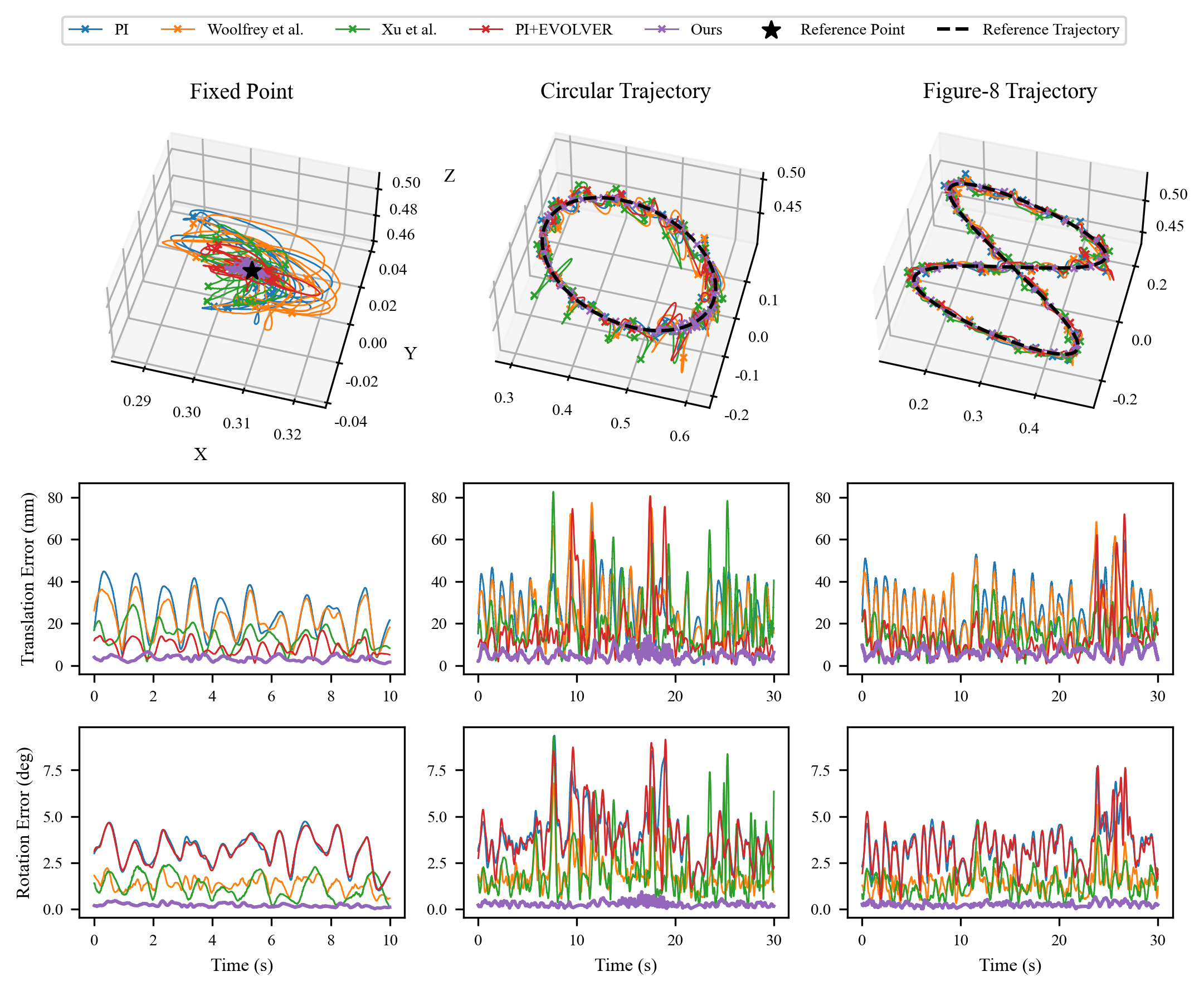}
    \caption{Simulation trajectory tracking results for the fixed-point, circular, and figure-8 tasks.
    The top row shows the end-effector trajectories in 3D space. The middle row shows the corresponding position errors over time for each method. The bottom row shows the corresponding rotation errors over time for each method.}
    \label{fig:trajectory}
\end{figure*}

The base motions are directly derived from recorded AUV motions subjected to wave disturbances in shallow-water conditions. The resulting base motion profile is shown in Fig. \ref{fig:sim_base_motion}. The base motion exhibits significant 6-DOF oscillations, with peak linear velocities reaching up to 0.32 m/s and angular velocities up to 24 deg/s.

\subsubsection{Baseline Controllers}

To evaluate the effectiveness of the proposed method, we compare it against four baseline controllers. 
These baselines are selected because they have been previously validated on real robotic manipulators:

\begin{itemize}
    \item \textbf{PI}. A standard kinematic feedback controller that does not rely on the system dynamics. It maps Cartesian errors to joint velocities via the pseudoinverse Jacobian:
    \begin{equation}
            \dot{\boldsymbol{q}}_{c}= \boldsymbol{J}^{\dagger}\rev{\boldsymbol{R}^{B}_{W}} \left[ \dot{\boldsymbol{x}}_{d}+ \boldsymbol{K}_{p}\boldsymbol{e}
            + \boldsymbol{K}_{i}\int \boldsymbol{e}dt \right],
    \end{equation}
    where \rev{$\boldsymbol{e}= \log \left( \boldsymbol{x}^{-1}\boldsymbol{x}_{d} \right)$}, $\boldsymbol{K}_{p} \in \mathbb{R}^{6\times6}$, and $\boldsymbol{K}_{i} \in \mathbb{R}^{6\times6}$ are the proportional and integral gain matrices, respectively.

   \item \textbf{Woolfrey et al.~\cite{peec}.} It employs an autoregressive (AR) model to predict base pose, effectively projecting the global reference into a dynamic local target. MPC is used to compute base frame velocities $\boldsymbol{u}$ to track the compensated reference. Then the desired joint velocities are computed via:
    $
        \dot{\boldsymbol{q}}_{c}= \boldsymbol{J}^{\dagger}(\boldsymbol{q}
        )\rev{\boldsymbol{u}_{0}^{*}}.
    $

    \item \textbf{Xu et al.~\cite{xuoe}}. It designs an estimator to estimate the joint velocity disturbances $\hat{\dot{\boldsymbol{q}}}_{\text{dist}}$ induced by base motion and compensate for them in the velocity command:
    \begin{equation}
            \dot{\boldsymbol{q}}_{c}= \boldsymbol{J}^{\dagger} \rev{\boldsymbol{u}_{0}^{*}} + \hat{\dot{\boldsymbol{q}}}_{\text{dist}},
    \end{equation}
    where \rev{$\boldsymbol{u}_{0}^{*}$} is the desired end-effector velocity computed by MPC.

    \item \textbf{PI+EVOLVER \cite{evolver}}. An enhanced kinematic controller that estimates the disturbance linear velocity of the base and compensates for it in the velocity command:
    \begin{equation}
            \dot{\boldsymbol{q}}_{c}= \boldsymbol{J}^{\dagger}\rev{\boldsymbol{R}^{B}_{W}} \left[ \dot{\boldsymbol{x}}_{d}+ \boldsymbol{K}_{p}\boldsymbol{e}
            + \boldsymbol{K}_{i}\int \boldsymbol{e}dt - \Delta \right],
    \end{equation}
    where $\Delta= [\hat{\boldsymbol{v}}_{\text{dist}}, \boldsymbol{0}]^T$. 
    EVOLVER is a Koopman operator-based observer that estimates $\hat{\boldsymbol{v}}_{\text{dist}}$ from historical end-effector pose measurements.

\end{itemize}

\rev{
The baseline parameters are tuned on the real manipulator based on the settings reported in the original papers. Detailed parameter settings and tuning procedures are provided in the supplementary material.
}
The parameters for our proposed method are listed in Table \ref{tab:params}. 

\begin{table}[htbp]
    \centering
    \caption{Parameters for the Proposed Method}
    \label{tab:params}
    \begin{tabular}{l}
        \toprule
        \textbf{Controller Parameters} \\
        \midrule
        $\boldsymbol{K}_p=\mathrm{diag}(500, 500, 500, 500, 500, 500)$ \\
        $\boldsymbol{K}_d=\mathrm{diag}(40, 40, 40, 40, 40, 40)$ \\
        $\boldsymbol{K}_{p, \text{ns}}=\mathrm{diag}(50,50,50,50,50,50,50)$\\
        $\boldsymbol{K}_{d, \text{ns}}=\mathrm{diag}(20,20,20,20,20,20,20)$\\
        $\lambda=0.001$ \\
        \midrule
        \textbf{Estimator Parameters} \\
        \midrule
        $
        \boldsymbol{Q}_d=\mathrm{diag}\!\left(
        \begin{aligned}
        &1\times10^{-9},\,1\times10^{-8},\,1\times10^{-4},\,1\times10^{-5},\\
        &1.5\times10^{-3},\,2\times10^{-3},\,1\times10^{-4},\\
        &1\times10^{-10},\,6\times10^{-6},\,1\times10^{-6}
        \end{aligned}
        \right)
        $ \\
        
        $\boldsymbol{R}_{\text{imu}}=\mathrm{diag}(3\times10^{-2},\,2\times10^{-4})$ \\

        $\boldsymbol{R}_{\text{pose}}=\mathrm{diag}(1\times10^{-8},\,1\times10^{-8},\,5\times10^{-4},\,5\times10^{-4})$ \\

         \bottomrule
    \end{tabular}
\end{table}

\subsubsection{Tracking Performance}
Tracking performance is shown in Fig. \ref{fig:trajectory} and summarized quantitatively in Table \ref{tab:exp1}. For position and rotation errors, we report the mean, standard deviation, and maximum values over the entire trajectory. In the Fixed Point task, the proposed method achieves a mean position error of 3.30 mm, representing reductions of 86.2\%, 84.9\%, 75.4\%, and 61.9\% relative to PI, \cite{peec}, \cite{xuoe}, and \cite{evolver}, respectively. The maximum position error is reduced to 6.54 mm, substantially lower than the baseline maximums. Similarly, the mean rotation error is 0.21 deg, corresponding to reductions of 93.6\%, 84.5\%, 82.9\%, and 93.5\% relative to PI, \cite{peec}, \cite{xuoe}, and \cite{evolver}, respectively. The low standard deviations for both position and rotation indicate that the proposed method effectively suppresses disturbances, achieving stable and precise end-effector control. In the Circular and Figure-8 tasks, the proposed method consistently maintains high tracking precision. In the Circular task, the mean position error is reduced to 6.14 mm, corresponding to \rev{reductions} of 77.5\%, 77.9\%, 70.6\%, and 61.6\% compared with PI, \cite{peec}, \cite{xuoe}, and \cite{evolver}, respectively. The mean rotation error remains below 0.7 deg, achieving reductions greater than 60\% relative to all baselines. Similarly, in the Figure-8 task, the proposed method achieves a mean position error of 6.36 mm, representing reductions of 75.8\%, 74.3\%, 57.8\%, and 53.1\% compared with the four baseline methods.

\begin{table}
    [htbp]
    \centering
    \caption{Simulation Trajectory Tracking Performance Metrics}
    \label{tab:exp1}
    \begin{tabular}{@{}llcccccc@{}}
        \toprule \multirow{2}{*}{\textbf{Task}} & \multirow{2}{*}{\textbf{Method}} & \multicolumn{3}{c}{\textbf{Position Error (mm)}} & \multicolumn{3}{c}{\textbf{Rotation Error (deg)}} \\
        \cmidrule(lr){3-5} \cmidrule(lr){6-8}   &                                  & \textbf{Mean}                                   & \textbf{Std.}                                    & \textbf{Max}    & \textbf{Mean}  & \textbf{Std.}  & \textbf{Max}   \\
        \midrule \multirow{5}{*}{\makecell{Fixed\\Point}}   &PI & 23.85 & 10.29 & 44.76 & 3.21 & 0.81 & 4.73 \\
&\cite{peec} & 21.90 & 8.39 & 37.99 & 1.33 & 0.36 & 2.22 \\
&\cite{xuoe} & 13.40 & 5.49 & 28.97 & 1.20 & 0.62 & 2.40 \\
&\cite{evolver} & 8.67 & 3.47 & 16.68 & 3.17 & 0.79 & 4.64 \\
&Ours & \textbf{3.30} & \textbf{1.23} & \textbf{6.54} & \textbf{0.21} & \textbf{0.09} & \textbf{0.46} \\
        \midrule \multirow{5}{*}{Circular}      &PI & 27.26 & 12.21 & 71.13 & 4.04 & 1.41 & 9.37 \\
&\cite{peec} & 27.74 & 13.47 & 77.42 & 1.95 & 1.03 & 6.79 \\
&\cite{xuoe} & 20.91 & 14.15 & 82.64 & 2.24 & 1.58 & 9.32 \\
&\cite{evolver} & 15.97 & 13.84 & 80.59 & 3.96 & 1.45 & 9.15 \\
&Ours & \textbf{6.14} & \textbf{3.03} & \textbf{17.47} & \textbf{0.7} & \textbf{0.36} & \textbf{1.67} \\
        \midrule \multirow{5}{*}{Figure-8}      &PI & 26.25 & 12.09 & 59.38 & 3.33 & 1.13 & 7.63 \\
&\cite{peec} & 24.78 & 11.56 & 68.30 & 1.41 & 0.70 & 5.64 \\
&\cite{xuoe} & 15.06 & 6.76 & 38.36 & 1.42 & 0.79 & 4.78 \\
&\cite{evolver} & 13.55 & 10.57 & 71.96 & 3.31 & 1.18 & 7.73 \\
&Ours & \textbf{6.36} & \textbf{2.67} & \textbf{13.26} & \textbf{0.25} & \textbf{0.12} & \textbf{0.65} \\
        \bottomrule
    \end{tabular}
\end{table}

{
\color{R1}

\subsection{Effects of Constraint Activation and Model Uncertainty}
\label{subsec:constraint_model_uncertainty}

The analysis in Section~\ref{subsec:impedance_analysis} identifies active inequality constraints and model uncertainty as two sources of deviation from the desired impedance dynamics, represented by $\boldsymbol{\delta}_c$ and $\boldsymbol{\delta}_d$, respectively.

\begin{figure}[htbp]
    \centering
    \includegraphics[width=\columnwidth]
    {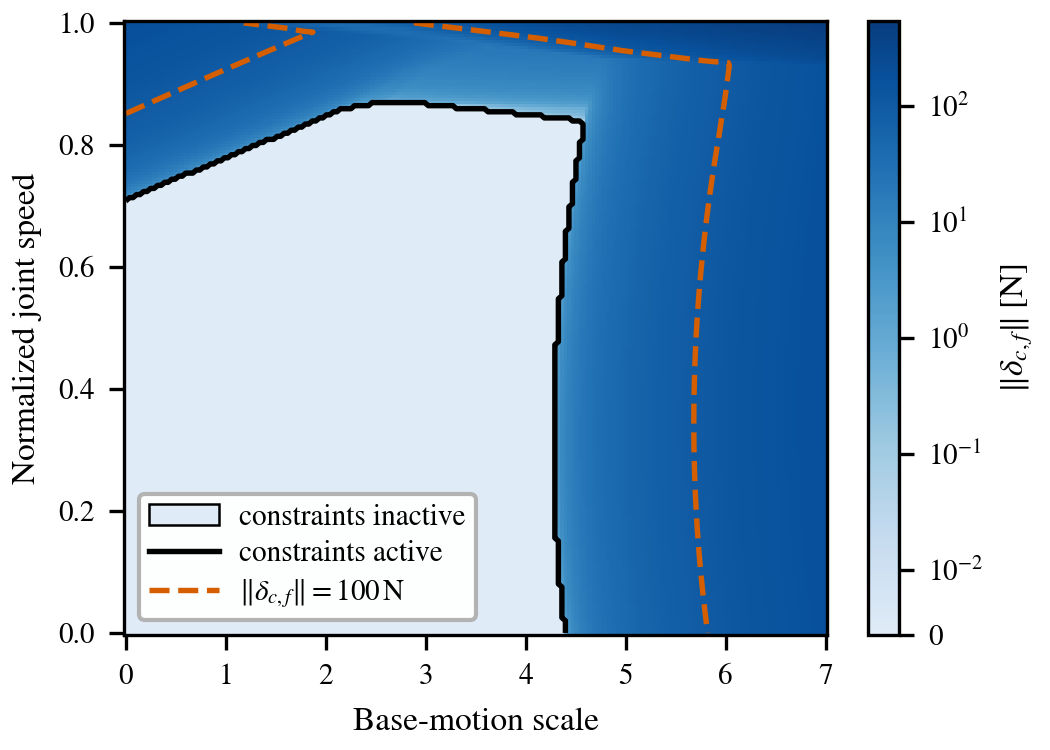}
    \caption{\rev{Constraint activation under different base-motion scales and normalized joint speeds. The contour denotes $\|\boldsymbol{\delta}_{c,f}\|=100~\mathrm{N}$, approximately the upper force limit of the Panda manipulator.}}
    \label{fig:constraint_wrench_contours}
\end{figure}

\subsubsection{Constraint Activation}

To quantify the effect of active inequality constraints on the realized task-space impedance, we perform a numerical parameter sweep over the normalized joint-speed level and the base-motion scaling factor. The normalized joint-speed level ranges from 0 to 1, corresponding to zero joint velocity and the respective joint-velocity limits. The base-motion scaling factor ranges from 0 to 7. At a base-motion scaling factor of 1, the prescribed linear and angular velocity vectors are
$[0.05,0.05,0.05]^T~\mathrm{m/s}$ and $[0.1,0.1,0.1]^T~\mathrm{rad/s}$, respectively, while the linear and angular acceleration vectors are $[0.2,0.2,0.2]^T~\mathrm{m/s^2}$ and $[0.3,0.3,0.3]^T~\mathrm{rad/s^2}$, respectively.
At each parameter combination, we solve the QP and compute the constraint-induced equivalent task-space force $\boldsymbol{\delta}_{c,f}$.

As shown in Fig.~\ref{fig:constraint_wrench_contours}, the inequality constraints remain inactive over a broad region of the normalized joint-speed and base-motion scaling-factor space. In this region, $\boldsymbol{\delta}_{c,f}=\boldsymbol{0}$, and the desired task-space impedance dynamics are realized without constraint-induced deviation.
Once the constraints become active, $\boldsymbol{\delta}_{c,f}$ becomes nonzero and generally increases in magnitude with the base-motion scaling factor, potentially resulting in larger tracking errors in free motion or higher interaction forces during contact.

\begin{figure}[tbp]
    \centering
    \includegraphics[width=\columnwidth]
    {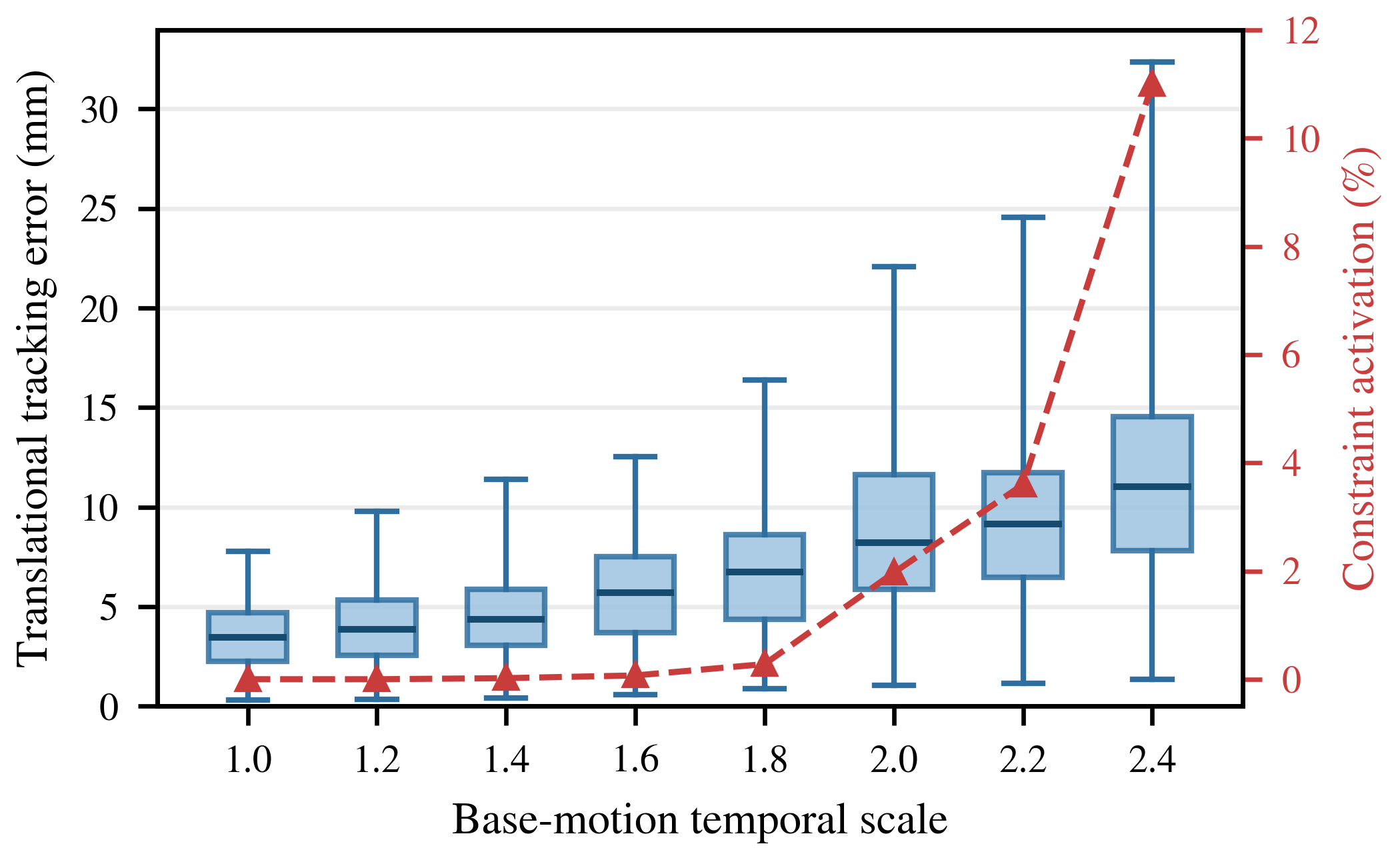}
    \caption{\rev{Translational tracking error and constraint-activation ratio under increasing base-motion temporal scales.}}
    \label{fig:constraint_tracking_error}
\end{figure}

\begin{figure}[tbp]
    \centering
    \includegraphics[width=\columnwidth]
    {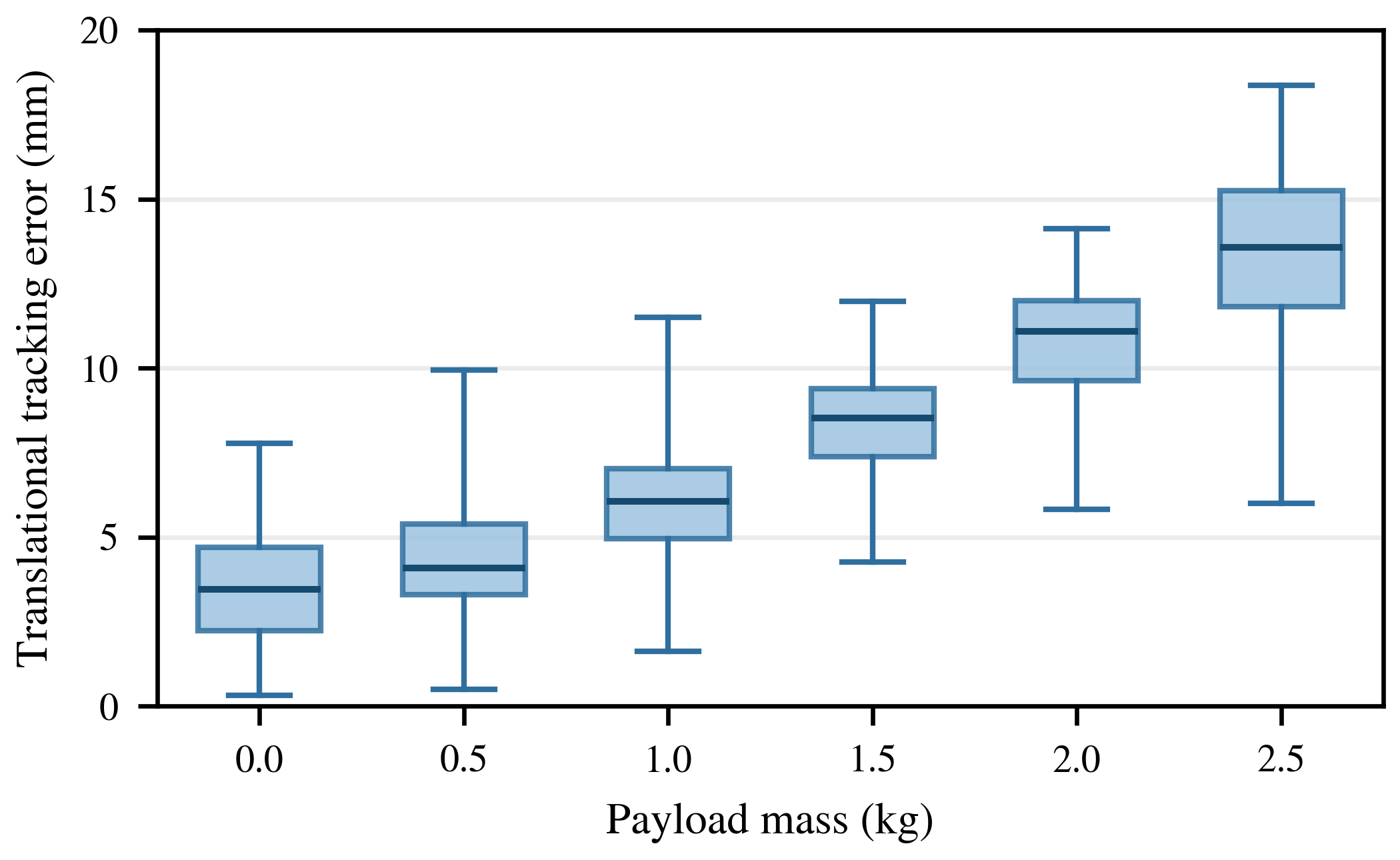}
    \caption{\rev{Translational tracking error under different unmodeled payloads.}}
    \label{fig:payload_tracking_error}
\end{figure}

\begin{figure*}[htbp]
    \centering
    \includegraphics[width=0.8\linewidth]{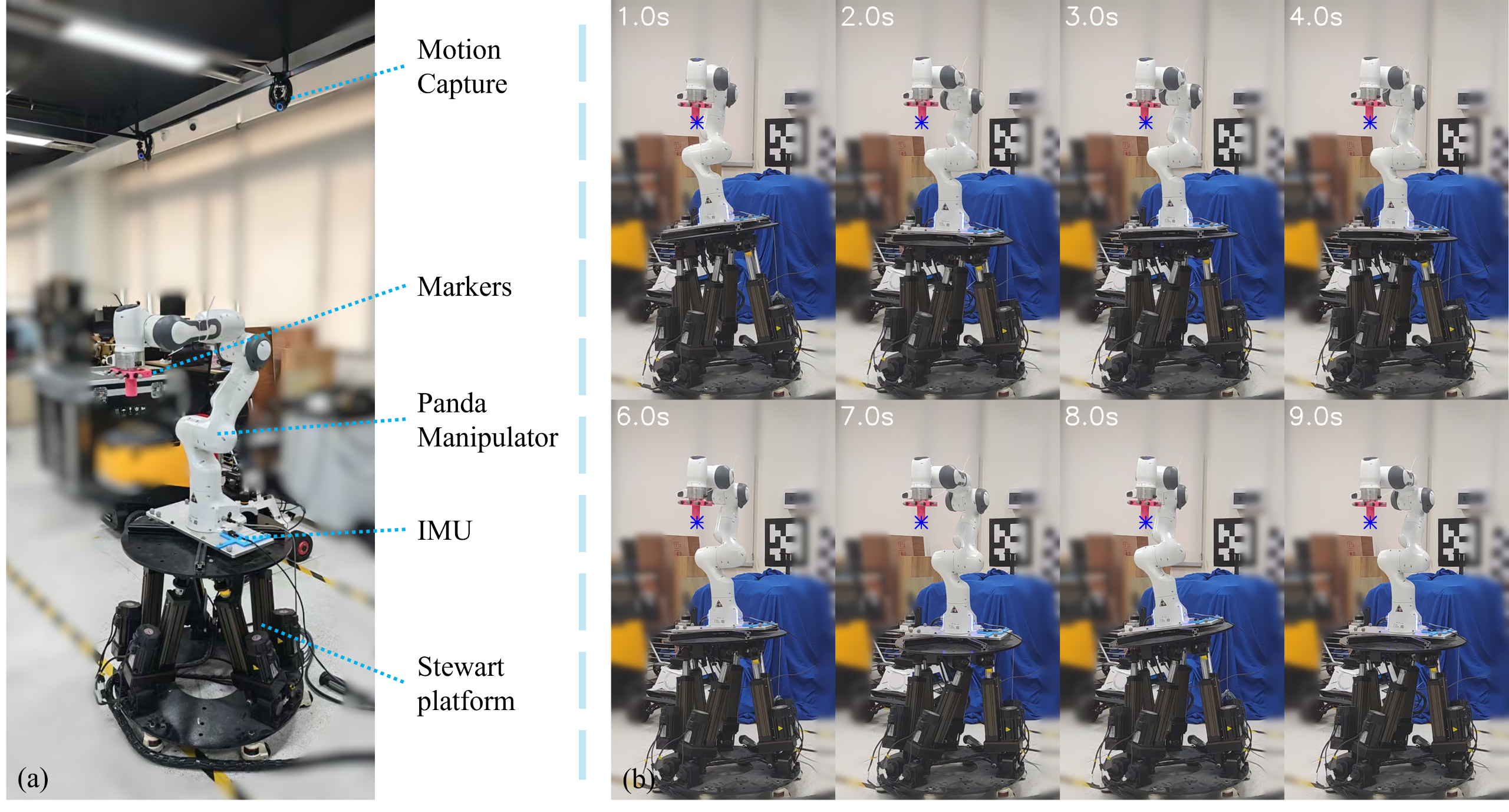}
    \caption{(a) Real-world experimental setup. (b) Snapshot of the real-world fixed-point stabilization experiment.}
    \label{fig:real_video}
\end{figure*}

To further evaluate the influence of constraint activation on tracking accuracy, Gazebo simulations are conducted with progressively increased base-motion frequency while keeping the trajectory amplitude fixed.

As shown in Fig.~\ref{fig:constraint_tracking_error}, the tracking error increases with the base-motion intensity because of stronger base-induced disturbances. 
No constraint is activated at motion scales of 1.0 and 1.2, whereas the activation ratio increases rapidly at higher scales.
At higher base-motion scales, the constraint-activation ratio increases, together with a more pronounced growth in the maximum tracking error.
Active constraints introduce a nonzero $\boldsymbol{\delta}_c$ into the error dynamics, which may degrade tracking performance and enlarge the ultimate tracking-error bound if its magnitude becomes large.

The influence of $\lambda$ is also evaluated at two representative base-motion scales. At scale 1.0, where the constraints remain inactive, changing $\lambda$ has almost no effect on the tracking accuracy. 
At scale 2.4, constraint activation makes the tracking performance sensitive to $\lambda$. Increasing $\lambda$ from $0.001$ to $10$ raises the maximum tracking error from $32.37~\mathrm{mm}$ to $92.08~\mathrm{mm}$, because the optimizer gives more weight to maintaining the nominal null-space posture and consequently permits a larger deviation from the primary task.

\subsubsection{Payload Uncertainty}

The effect of model uncertainty is evaluated through Gazebo simulations with unmodeled payloads attached to the end-effector. The additional payload is varied from $0$ to $2.5~\mathrm{kg}$, while the controller retains the nominal dynamic model. The nominal end-effector mass is $0.73~\mathrm{kg}$.

As shown in Fig.~\ref{fig:payload_tracking_error}, the translational tracking error increases with the unmodeled payload. 
The error distribution remains relatively concentrated, indicating that the payload mismatch mainly introduces an additional steady-state tracking error rather than large oscillations.

Overall, the two disturbance terms affect the closed-loop behavior in different ways. In the payload simulations, the model uncertainty is approximately constant and therefore mainly contributes to $\boldsymbol{\delta}_d$ as a slowly varying disturbance, resulting primarily in additional steady-state tracking errors. 
In contrast, $\boldsymbol{\delta}_c$ may vary rapidly when inequality constraints are activated, released, or switched. It therefore has a stronger influence on the transient response and may lead to larger variations in the tracking error.

\subsection{Timing Performance}

To evaluate the real-time performance of the proposed controller, a timing benchmark is performed on two CPU platforms. The benchmark measures the computation time for one control cycle, including state update, forward kinematics computation, QP problem construction, QP solving, and solution decoding. For each platform, $10^6$ random cases are generated. The joint positions and nominal-posture reference configurations are uniformly sampled within the joint position limits, and the joint velocities are sampled within the velocity limits. The base translational coordinates and the roll, pitch, and yaw angles are sampled within $\pm 0.5~\mathrm{m}$ and $\pm 0.5~\mathrm{rad}$, respectively. The base linear/angular velocities and accelerations are sampled within $\pm 10~\mathrm{m/s}$, $\pm 10~\mathrm{rad/s}$, $\pm 10~\mathrm{m/s^2}$, and $\pm 10~\mathrm{rad/s^2}$, respectively. The desired end-effector pose is generated by adding random position and orientation offsets within $\pm 0.5~\mathrm{m}$ and $\pm 0.5~\mathrm{rad}$ to the current end-effector pose.

\begin{table}[htbp]
    \centering
    \caption{\rev{Timing Performance of the Proposed Controller}}
    \label{tab:timing_performance}
    \begin{tabular}{lcccc}
        \toprule
        CPU & Total avg. & Total max. & QP avg. & QP max. \\
            & (ms) & (ms) & (ms) & (ms) \\
        \midrule
        i9-13900K & 0.0161 & 0.0859 & 0.0039 & 0.0733 \\
        i7-13620H & 0.0212 & 0.1885 & 0.0051 & 0.0856 \\
        \bottomrule
    \end{tabular}
\end{table}

Table~\ref{tab:timing_performance} reports the timing results. The mean total computation time is $0.0161~\mathrm{ms}$ on the i9-13900K and $0.0212~\mathrm{ms}$ on the i7-13620H. The maximum total computation time is $0.0859~\mathrm{ms}$ and $0.1885~\mathrm{ms}$, respectively. The mean QP solution time is $0.0039~\mathrm{ms}$ on the i9-13900K and $0.0051~\mathrm{ms}$ on the i7-13620H, accounting for a small fraction of the total computation time.

Among the $10^6$ random samples, $8901$ cases are infeasible, mainly due to joint torque limit violations caused by large sampled base motions. Since the computation time does not exceed the $1~\mathrm{ms}$ control period for any sampled case, the results indicate that the proposed controller satisfies the timing requirement of the $1~\mathrm{kHz}$ control loop.
}

\subsection{Real-world Trajectory Tracking Experiments}

\begin{figure}[htbp]
    \centering
    \includegraphics[width=\linewidth]{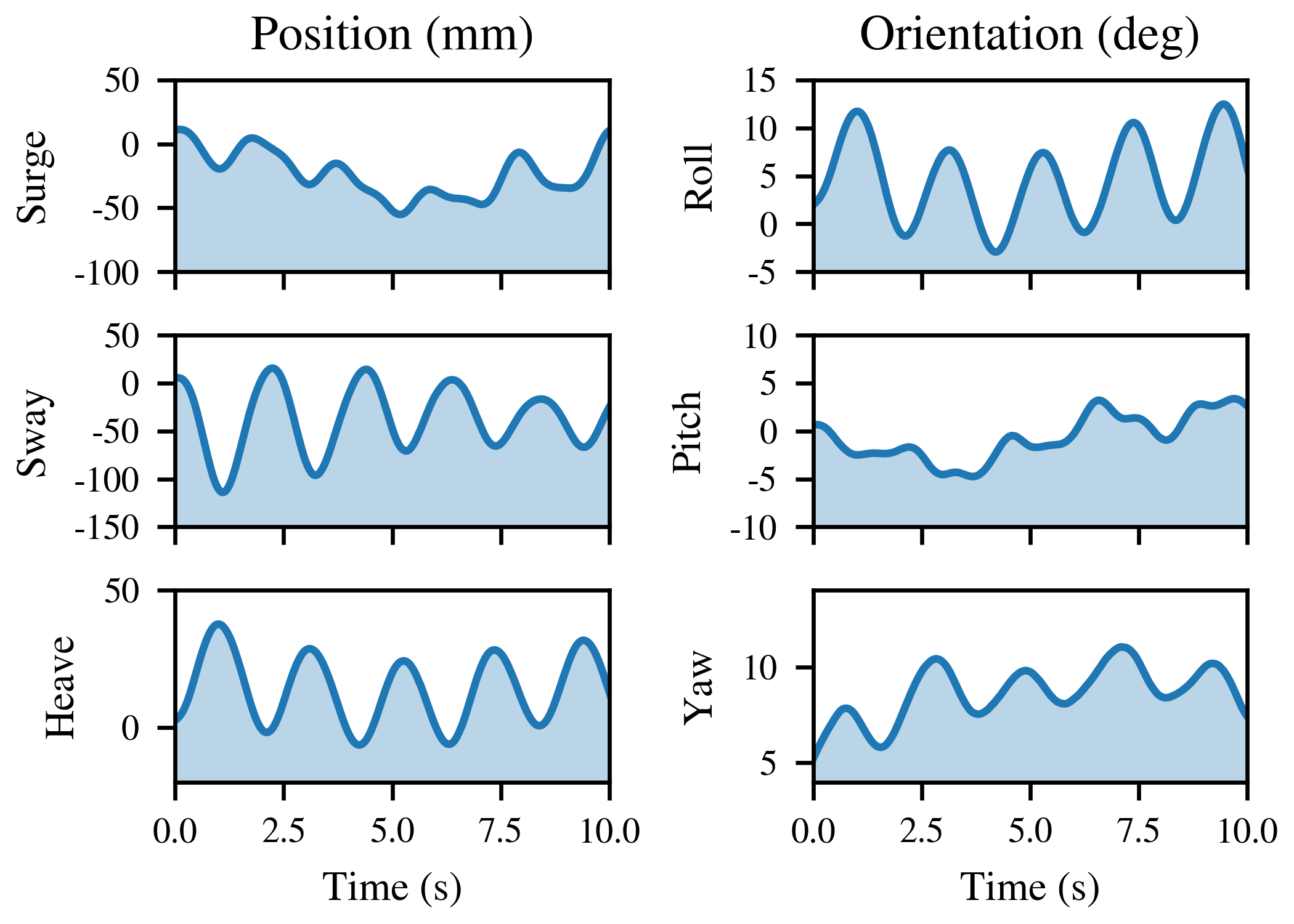}
    \caption{Base motion for real-world experiments.}
    \label{fig:real_base_motion}
\end{figure}

\begin{figure*}[tbp]
    \centering
    \includegraphics[width=\linewidth]{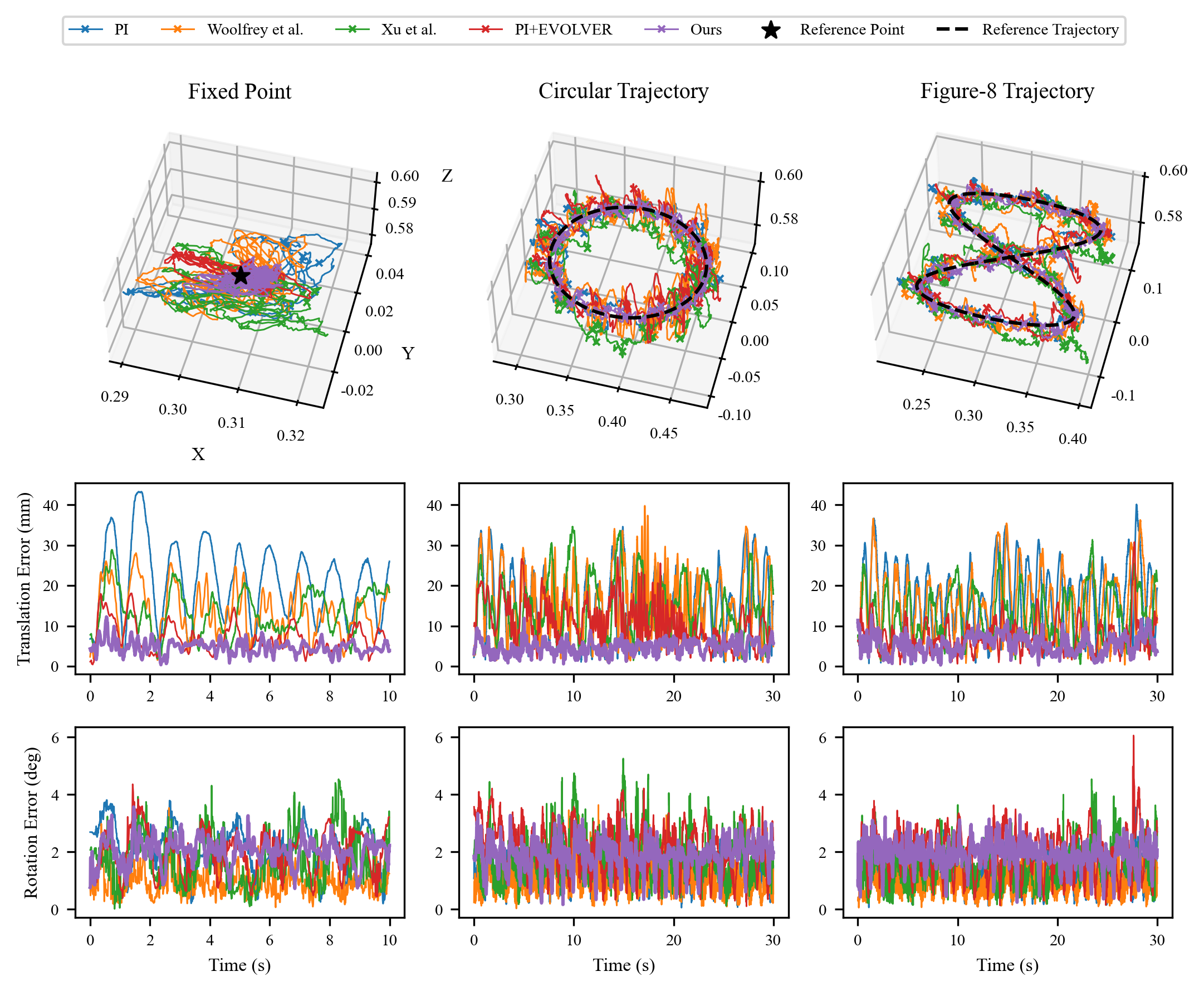}
    \caption{Real-world trajectory tracking results for the fixed-point, circular, and figure-8 tasks.}
    \label{fig:real_trajectory}
\end{figure*}

\subsubsection{Hardware Setup}
To validate the proposed method in real-world scenarios, experiments are conducted on a ship-motion simulation platform.
As shown in Fig.~\ref{fig:real_video}(a), the experimental setup consists of a 7-DOF Franka Emika Panda manipulator mounted on a 6-DOF Stewart platform. The platform is actuated by six high-precision electric cylinders and can reproduce 6-DOF ship motions.
A HiPNUC13R2 IMU is rigidly mounted on the platform to measure base angular velocity and linear acceleration at 100~Hz. An external optical motion capture system tracks markers attached to both the manipulator end-effector and the platform, providing ground-truth poses of the end-effector and the base expressed in the inertial frame at 120~Hz.
All control algorithms are executed on a laptop equipped with an Intel i7-\rev{13620H} CPU. The IMU communicates with the control computer via USB, while the motion capture system, the manipulator, and the Stewart platform interface through EtherCAT. The manipulator is operated using the Franka ROS driver in torque control mode at a control frequency of 1~kHz. The Stewart platform is controlled at 100~Hz, with a closed-loop PID controller running on its onboard controller to track the desired platform trajectories.

\subsubsection{Task Description and Base Motion}

Due to the physical workspace limits of the Stewart platform and the motion capture's field of view, both the base motion amplitudes and the tracking trajectories are scaled down relative to the simulation. 
As shown in Fig. \ref{fig:real_base_motion}, the frequency characteristics and stochastic properties of the wave-induced motion are preserved. The peak linear and angular velocities reach 0.20 m/s and 20 deg/s, respectively. 
The circular trajectory has a radius of 0.075 m, and the figure-8 trajectory is defined by $x = 0.075\sin(t)$ and $y = 0.1125\sin(t)\cos(t)$.

\begin{table}
    [htbp]
    \centering
    \caption{Real-world Trajectory Tracking Performance Metrics}
    \label{tab:exp_real}
    \begin{tabular}{@{}llcccccc@{}}
        \toprule \multirow{2}{*}{\textbf{Task}} & \multirow{2}{*}{\textbf{Method}} & \multicolumn{3}{c}{\textbf{Position Error (mm)}} & \multicolumn{3}{c}{\textbf{Rotation Error (deg)}} \\
        \cmidrule(lr){3-5} \cmidrule(lr){6-8}   &                                  & \textbf{Mean}                                   & \textbf{Std.}                                    & \textbf{Max}    & \textbf{Mean}  & \textbf{Std.}  & \textbf{Max}   \\
        \midrule \multirow{5}{*}{\makecell{Fixed\\Point}}   &PI & 21.09 & 9.31 & 43.29 & 2.13 & 0.78 & 3.81 \\
&\cite{peec} & 13.41 & 6.07 & 28.08 & \textbf{0.91} & \textbf{0.44} & \textbf{3.54} \\
&\cite{xuoe} & 13.53 & 5.38 & 28.88 & 1.72 & 0.90 & 4.54 \\
&\cite{evolver} & 7.03 & 3.63 & 18.08 & 2.01 & 0.75 & 4.36 \\
&Ours & \textbf{4.73} & \textbf{1.67} & \textbf{12.27} & 2.02 & 0.46 & 3.57 \\
        \midrule \multirow{5}{*}{Circular}      &PI & 16.85 & 7.94 & 34.62 & 1.33 & \textbf{0.51} & \textbf{3.11} \\
&\cite{peec} & 16.33 & 7.75 & 39.79 & \textbf{1.08} & 0.53 & 3.64 \\
&\cite{xuoe} & 15.12 & 7.28 & 34.64 & 1.79 & 0.84 & 5.26 \\
&\cite{evolver} & 11.10 & 5.19 & 27.07 & 2.15 & 0.82 & 4.22 \\
&Ours & \textbf{4.70} & \textbf{1.65} & \textbf{11.91} & 1.87 & \textbf{0.51} & 3.34 \\
        \midrule \multirow{5}{*}{Figure-8}      &PI & 17.22 & 8.42 & 40.13 & 1.28 & 0.50 & 3.55 \\
&\cite{peec} & 16.03 & 7.69 & 36.64 & \textbf{0.89} & \textbf{0.41} & \textbf{2.81} \\
&\cite{xuoe} & 13.11 & 6.10 & 31.37 & 1.46 & 0.67 & 4.54 \\
&\cite{evolver} & 7.31 & 3.87 & 30.79 & 1.95 & 0.78 & 6.06 \\
&Ours & \textbf{5.43} & \textbf{2.32} & \textbf{12.74} & 1.90 & 0.46 & 3.30 \\
        \bottomrule
    \end{tabular}
\end{table}

\subsubsection{Tracking Performance}
The real-world tracking results are shown in Fig. \ref{fig:real_trajectory}, and quantitative metrics are summarized in Table \ref{tab:exp_real}. The proposed method consistently improves position tracking accuracy across all tasks. In the Fixed-Point experiment, the mean position error is 4.73 mm, corresponding to reductions of 77.6\%, 64.7\%, 65.0\%, and 32.7\% relative to PI, \cite{peec}, \cite{xuoe}, and \cite{evolver}, respectively. The maximum error is reduced to 12.27 mm, yielding improvements of 71.7\%, 56.3\%, 57.5\%, and 32.2\%. For the Circular trajectory, the mean error decreases to 4.70 mm, representing reductions of 72.1\%, 71.2\%, 68.9\%, and 57.7\% compared with the baselines. The maximum error is limited to 11.91 mm, substantially lower than the baselines. In the Figure-8 task, the proposed method achieves a mean error of 5.43 mm, reducing errors by 68.5\%, 66.1\%, 58.6\%, and 25.7\%, respectively. 

All evaluated controllers achieve adequate orientation tracking with mean errors below 2.5 deg. The mean and maximum rotation errors of the proposed method are not uniformly lower than \rev{those of} all baselines.
As shown in Fig. \ref{fig:real_trajectory}, minor steady-state orientation errors occur. This behavior is mainly due to inaccuracies in real-system modeling, such as friction and parameter uncertainty.

Compared with simulation, the relative improvements in real-world experiments are smaller. This is due to (1) model mismatch and sensor/actuator nonidealities in the real system, and (2) smaller base-motion amplitudes in real-world experiments. In simulation, larger base disturbances induce stronger inertial effects, where the proposed method becomes more advantageous.

\subsection{Base State Estimation Experiments}

\begin{table*}[ht]
    \centering
    \caption{\rev{RMSE of the base-state estimates in the ablation experiments.}}
    \label{tab:estimator_ablation}
    \renewcommand{\arraystretch}{1.18}
    \setlength{\tabcolsep}{5pt}

    \resizebox{\textwidth}{!}{%
    \begin{tabular}{lcccccc}
        \toprule
        Method
        & \multicolumn{2}{c}{Pose RMSE}
        & \multicolumn{4}{c}{Motion-State RMSE} \\
        \cmidrule(lr){2-3}
        \cmidrule(lr){4-7}
        & Position
        & Orientation
        & Linear Velocity
        & Angular Velocity
        & Linear Acceleration
        & Angular Acceleration \\
        & (mm)
        & (deg)
        & ($\mathrm{m/s}$)
        & ($\mathrm{rad/s}$)
        & ($\mathrm{m/s^2}$)
        & ($\mathrm{rad/s^2}$) \\
        \midrule

        Direct base pose + IMU
        & 0.2307 & 0.06159
        & 0.009794 & 0.03007
        & 0.1906 & 0.5442 \\

        Direct base and end-effector poses + IMU
        & 0.2087 & 0.05716
        & 0.009408 & 0.03001
        & 0.1851 & 0.5367 \\

        Direct base pose only
        & 0.2656 & 0.06370
        & 0.01186 & 0.04299
        & 0.2184 & 0.7132 \\

        \midrule

        Proposed w/o FK-derived base pose
        & 10.52 & 2.952
        & 0.01679 & 0.03484
        & 0.2041 & 0.5274 \\

        Proposed w/o end-effector pose
        & 7.993 & 0.5964
        & 0.04237 & 0.03348
        & 0.2150 & 0.5274 \\

        Proposed w/o IMU
        & 7.817 & 0.8316
        & 0.05003 & 0.09643
        & 0.3227 & 0.6772 \\

        Reduced-state ESKF
        & 6.048 & 0.6314
        & 0.02009 & 0.03456
        & 0.2115 & 0.5330 \\

        Proposed full
        & 6.013 & 0.6313
        & 0.01687 & 0.03485
        & 0.2035 & 0.5274 \\

        \bottomrule
    \end{tabular}%
    }
\end{table*}

\begin{figure*}[htbp]
    \centering
    \includegraphics[width=\linewidth]{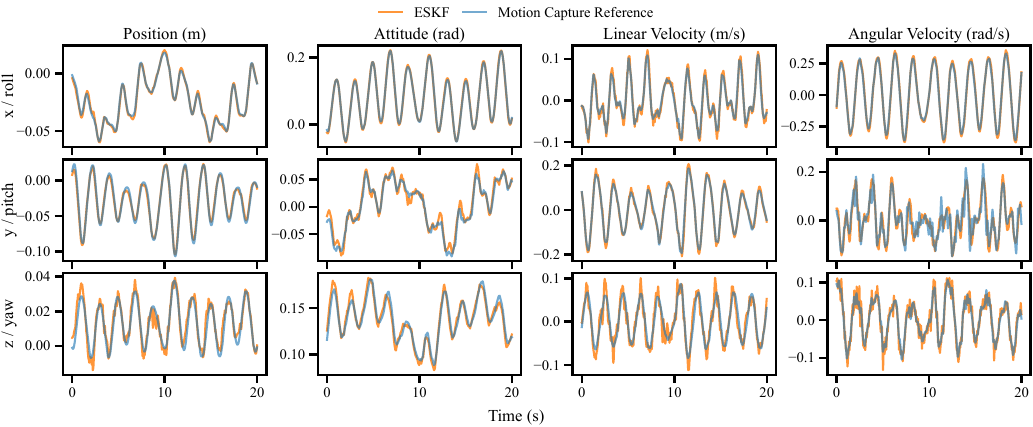}
    \caption{\rev{Base pose and velocity estimates compared with the
    motion-capture references.}}
    \label{fig:eskf_pose_velocity}
\end{figure*}

\begin{figure}[htbp]
    \centering
    \includegraphics[width=\linewidth]{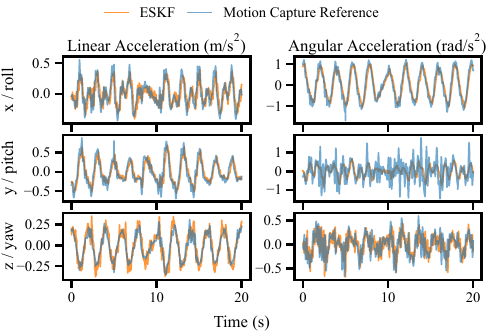}
    \caption{\rev{Base acceleration estimates compared with the
    motion-capture references.}}
    \label{fig:eskf_acceleration}
\end{figure}

\rev{
To quantitatively evaluate the proposed base-state estimator, the base pose measured by the motion capture system is used as the reference. 
A constant rigid-body transformation exists between the marker frame attached to the Stewart platform and the robot base frame. 
This transformation is calibrated using the measured end-effector pose, the joint configuration, and the manipulator kinematic model, and is then kept fixed throughout all experiments.
The calibrated marker measurements are transformed into the robot base frame to obtain the reference base position and orientation.
The reference linear and angular velocities are obtained by differentiating the base pose trajectories and filtering the results to reduce noise. The reference linear and angular accelerations are obtained through further differentiation and filtering.
}

{
\color{R1}

Ablation experiments are conducted to evaluate the effects of different measurements and augmented motion states on estimation performance.
Eight configurations are evaluated: three configurations using direct base-pose measurements, namely base pose with IMU, base and end-effector poses with IMU, and base pose alone; three ablated configurations obtained by removing the FK-derived base-pose measurement, the end-effector pose measurement, or the IMU measurement from the proposed estimator; the proposed estimator; and a reduced-state ESKF. 
The reduced-state ESKF omits $\boldsymbol{\omega}_{B}^{B}$, $\boldsymbol{a}_{B}^{B}$, and $\boldsymbol{\alpha}_{B}^{B}$ from the state vector. 
Since these states are still required by the proposed controller, angular velocity and angular acceleration are estimated using an auxiliary Kalman filter, while linear acceleration is obtained by filtering the accelerometer measurements.

Figures~\ref{fig:eskf_pose_velocity} and \ref{fig:eskf_acceleration} compare the estimated base states with the motion-capture references. 
The estimated velocities agree closely with the references, while the estimated accelerations reproduce the main variations. 
Due to kinematic-model errors, the estimated base pose occasionally exhibits small biases. Nevertheless, the errors remain within an acceptable range, indicating reliable reconstruction of the base motion.

The quantitative results are summarized in Table~\ref{tab:estimator_ablation}. 
Removing the IMU causes the most significant degradation in the velocity and acceleration estimates, confirming its importance for reconstructing rapidly varying base motion. In contrast, removing the FK-derived base-pose measurement mainly degrades the position and orientation estimates, showing that this pseudo-measurement provides an absolute pose constraint and limits long-term drift. 
Removing the end-effector pose measurement has a stronger effect on the translational states, especially the linear velocity estimate. 
The angular estimates are less affected by the pose-measurement ablations because the gyroscope directly measures the base angular velocity.

The proposed estimator achieves slightly lower estimation errors than the reduced-state ESKF. It also directly estimates all base states required by the proposed controller, whereas the reduced-state formulation requires additional filtering to recover the omitted angular-velocity and acceleration states. 
Direct incorporation of an accurate base-pose measurement yields the lowest estimation errors. Moreover, adding the end-effector pose measurement to the direct base-pose and IMU measurements further reduces all RMSE values, indicating that it provides additional information for base-state estimation.
Thus, the proposed fusion framework can also improve base-state estimation when direct base-pose measurements are available. When such measurements are unavailable, the estimator can reconstruct the base state using the base-mounted IMU and end-effector pose feedback.
}

\subsection{Dynamic Peg-in-Hole Insertion Experiment}

\begin{figure*}[htbp]
    \centering
    \includegraphics[width=\linewidth]{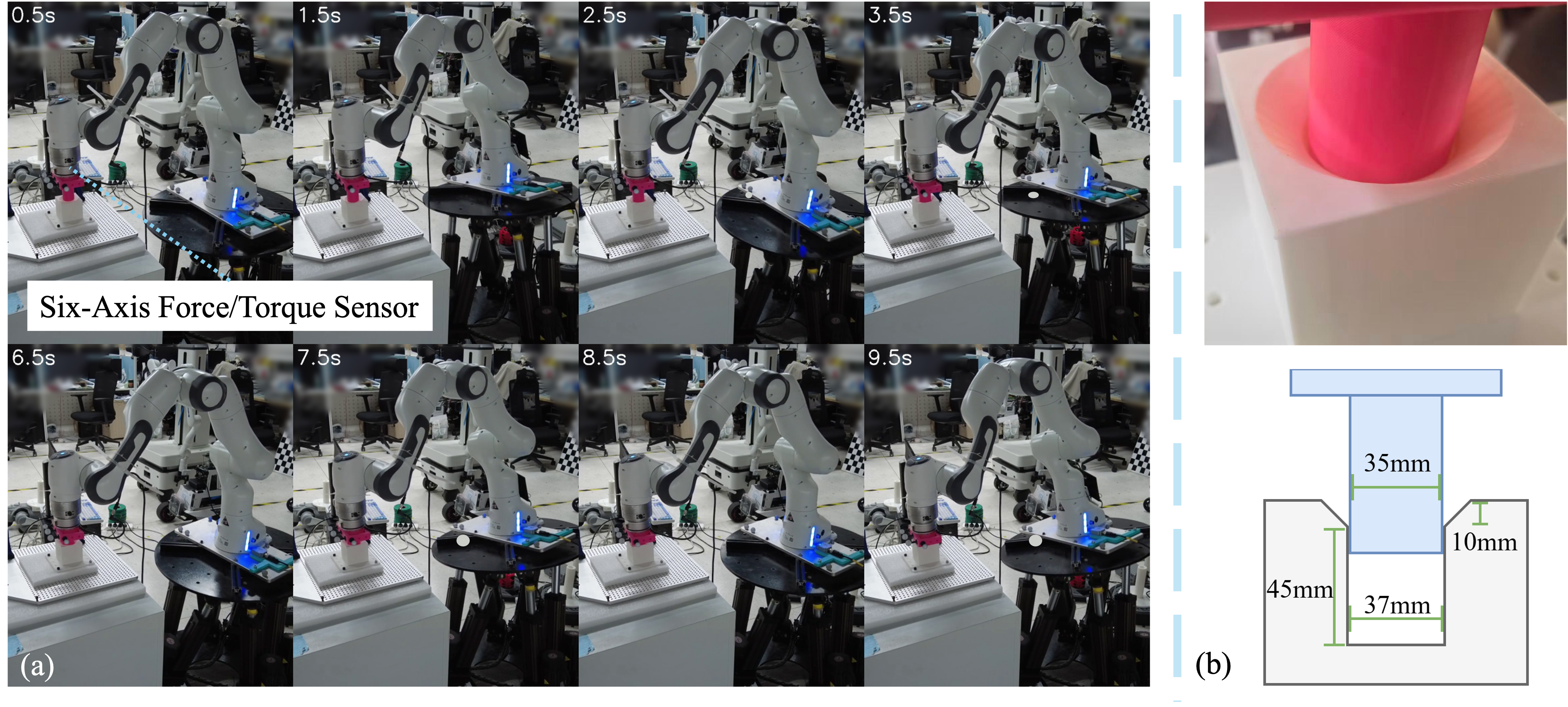}
    \caption{(a) Snapshot of the peg-in-hole insertion experiment. (b) Geometric configuration of the peg and hole.}
    \label{fig:peginhole}
\end{figure*}

\begin{figure}[htbp]
    \centering
    \includegraphics[width=\linewidth]{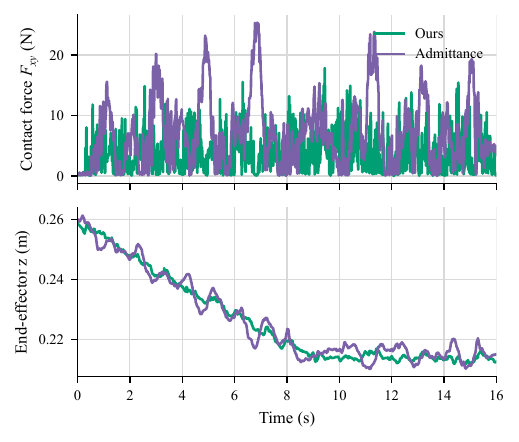}
    \caption{\rev{Contact force $F_{xy}$ and insertion depth $z$ during peg-in-hole insertion at a base-motion scale of 1.0 for the admittance controller and the proposed method.}}
    \label{fig:force}
\end{figure}

\begin{figure}[htbp]
    \centering
    \includegraphics[width=\linewidth]{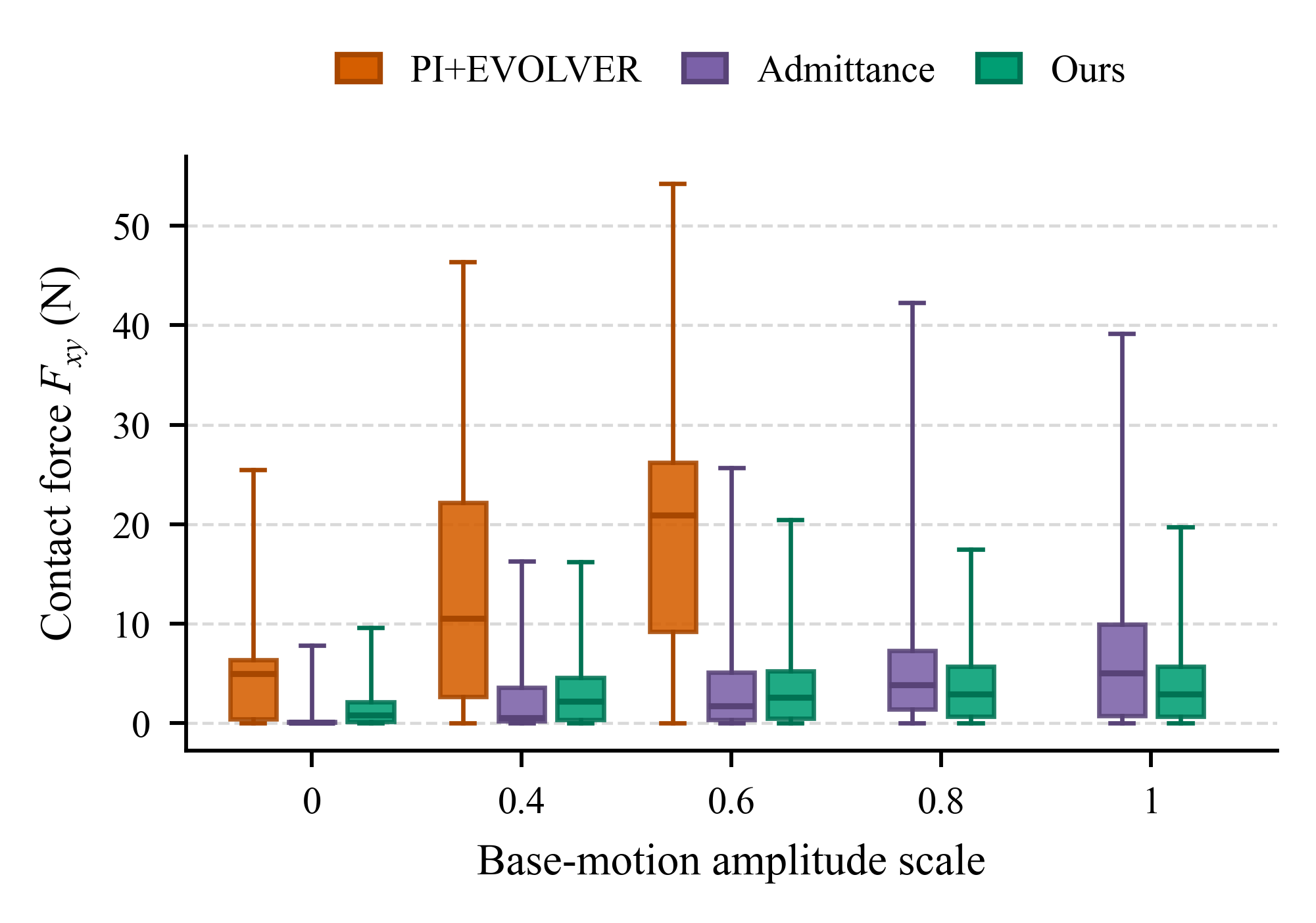}
    \caption{\rev{Box plots of $F_{xy}$ during successful peg-in-hole insertions under different base-motion scales.}}
    \label{fig:force_scale}
\end{figure}

To evaluate the proposed method in contact-rich scenarios, a dynamic peg-in-hole insertion experiment is conducted. The task requires the manipulator to insert a peg into a \rev{stationary hole secured by a bar clamp} while the robot base undergoes 6-DOF motion.

\subsubsection{Experimental Setup and Baselines}

As shown in Fig. \ref{fig:peginhole}, the peg has a diameter of 35 mm, while the hole has a diameter of 37 mm and a depth of 45 mm, resulting in a radial clearance of only 1 mm. A 10-mm chamfer is added to the hole entrance to provide passive guidance. The manipulator is commanded to move downward at a low velocity to complete the insertion. This narrow clearance dictates that the controller must exhibit sufficient compliance; otherwise, base motion will cause excessive contact wrenches or mechanical jamming. {\color{R1}A six-axis force/torque sensor is mounted between the manipulator wrist and the peg to measure the interaction wrench and provide feedback for the admittance controller.

The proposed method is compared with PI+EVOLVER~\cite{evolver} and an admittance controller. PI+EVOLVER is used as the best-performing kinematic baseline, while the admittance controller serves as a compliant baseline. 
Two admittance-control implementations are evaluated in preliminary experiments. The first uses an outer-loop admittance model to generate a Cartesian pose offset, which is tracked by PI+EVOLVER. The second adds a force-feedback velocity correction directly to the Cartesian velocity command generated by PI+EVOLVER, following~\cite{vasiljevic2025robust}. 
The first implementation provides more stable insertion behavior than the second in our setup and is therefore adopted as the admittance baseline.

The admittance offset $\boldsymbol{x}_{a}$ is generated by
\begin{equation}
    \boldsymbol{M}_{a}\ddot{\boldsymbol{x}}_{a}
    +\boldsymbol{D}_{a}\dot{\boldsymbol{x}}_{a}
    +\boldsymbol{K}_{a}\boldsymbol{x}_{a}
    =\boldsymbol{F}_{m},
    \label{eq:admittance_baseline}
\end{equation}
where $\boldsymbol{F}_{m}$ is the measured wrench. The admittance dynamics are implemented only along the $x$- and $y$-axes and about the roll and pitch axes. The corresponding parameter matrices are
$\boldsymbol{M}_{a}=\operatorname{diag}(8,8,0.02,0.02)$,
$\boldsymbol{D}_{a}=\operatorname{diag}(160,160,0.8,0.8)$, and
$\boldsymbol{K}_{a}=\operatorname{diag}(4000,4000,30,30)$.
A deadband of $1.5~\mathrm{N}$ is applied to the force measurements, and a deadband of $0.03~\mathrm{N\,m}$ is applied to the torque measurements. 

To evaluate the effect of base-motion amplitude, the 6-DOF platform trajectory is multiplied by scale factors of $0$, $0.4$, $0.6$, $0.8$, and $1.0$. Five trials are performed for each controller at scales $0$, $0.4$, $0.6$, and $0.8$. PI+EVOLVER fails in all trials at scale $0.8$ and is therefore not further tested at scale $1.0$. 
At scale $1.0$, the proposed and admittance controllers are each evaluated over 15 trials. An insertion is regarded as successful when the peg reaches the required insertion depth without jamming.

\subsubsection{Experimental Results}

Because misalignment in the plane perpendicular to the insertion direction is the main cause of jamming, contact performance is evaluated using $F_{xy}=\sqrt{F_x^2+F_y^2}$. 
Fig.~\ref{fig:force} shows representative force profiles during insertion, while Fig.~\ref{fig:force_scale} summarizes the distributions of $F_{xy}$ over repeated trials. Only successful trials are included in the force statistics.

As the base-motion amplitude increases, $F_{xy}$ generally increases for all controllers. PI+EVOLVER produces substantially larger forces than the two compliant controllers. At scale $0.6$, its mean and maximum $F_{xy}$ values reach $18.081~\mathrm{N}$ and $54.221~\mathrm{N}$, respectively, compared with $3.253~\mathrm{N}$ and $20.431~\mathrm{N}$ for the proposed method.
At scale $0.8$, PI+EVOLVER fails in all trials because the peg jams during insertion, and the resulting excessive contact force triggers the manipulator's protective stop.

As shown in Fig.~\ref{fig:force_scale}, the admittance controller achieves a lower mean $F_{xy}$ under small base motion. At scales $0$ and $0.4$, its mean $F_{xy}$ values are $0.258~\mathrm{N}$ and $2.197~\mathrm{N}$, respectively, compared with $1.265~\mathrm{N}$ and $2.875~\mathrm{N}$ for the proposed method. 
The proposed method becomes more advantageous as the base-motion amplitude increases. At scale $0.8$, it reduces the mean $F_{xy}$ from $5.316~\mathrm{N}$ to $3.572~\mathrm{N}$ and the maximum $F_{xy}$ from $42.271~\mathrm{N}$ to $17.454~\mathrm{N}$, corresponding to reductions of $32.8\%$ and $58.7\%$, respectively.  At scale $1.0$, the mean $F_{xy}$ is reduced from $6.588~\mathrm{N}$ to $3.619~\mathrm{N}$, corresponding to a reduction of $45.1\%$. The maximum $F_{xy}$ is reduced from $39.148~\mathrm{N}$ to $19.751~\mathrm{N}$, corresponding to a reduction of $49.5\%$.
The proposed method succeeds in all 15 trials (100\%), whereas the admittance controller succeeds in 12 of 15 trials (80\%).

The admittance controller modifies the Cartesian reference according to the measured wrench, and the modified reference is then tracked by the inner velocity controller. Under large base motion, this cascaded structure is less effective in dynamic contact than the proposed torque-level impedance controller. 
In contrast, the proposed method compensates for the base-induced dynamics and realizes task-space impedance directly at the torque level. It therefore maintains lower contact forces and achieves more reliable insertion as the base-motion amplitude increases.
}

%% file: conclusion.tex
\section{Discussion}
\subsection{\rev{Model Uncertainty and Constraint Activation}}

Due to modeling inaccuracies, the proposed method exhibits moderate steady-state errors in real-world experiments. These errors are primarily attributed to unmodeled friction, parameter uncertainty, and sensor noise. Although the disturbance compensation mechanism effectively mitigates base-induced inertial effects, residual model errors limit the achievable steady-state accuracy. \rev{
In our experiments, the controller remains effective with a $0.3~\mathrm{kg}$ unmodeled payload introduced by the force/torque sensor mounted at the end-effector, and successfully completes the peg-in-hole task. For substantially larger payloads, online payload identification or adaptive compensation may be required to maintain tracking and interaction performance.
}

\rev{
When the base motion becomes large, constraints associated with joint or torque limits may become active. The constrained QP may then fail to realize the desired impedance dynamics exactly. 
Unlike the relatively slow effects of model uncertainty, the constraint-induced correction may change abruptly when the QP active set changes.
This can degrade transient performance and increase tracking errors and contact-force variations. Although the QP keeps the commanded accelerations and torques within prescribed limits, tracking accuracy and compliance may be reduced under strong or frequent constraint activation. Therefore, trajectory planning that accounts for joint and torque constraints may help reduce
these effects.
}

\subsection{Perception and State Estimation}

\rev{
The proposed estimator requires the end-effector pose expressed in an inertial frame. In the present experiments, this measurement is provided by a motion capture system. This setup isolates the estimation and control performance from errors introduced by a particular onboard perception system, but reduces the practical realism of the experiments.
}

\rev{
The proposed framework mainly considers an eye-in-hand configuration, where end-effector pose feedback can be obtained by observing a stationary target or environmental landmark that defines a local inertial frame. The results also show that incorporating end-effector pose feedback with base-pose and IMU measurements improves estimation accuracy. For moving targets, the ESKF may be extended by augmenting the target pose and velocity states and incorporating relative visual measurements together with GNSS measurements. The effects of visual noise and delay are not considered in this work and will be investigated in future experiments with onboard perception.
}

\subsection{\rev{Manipulator-Induced Ship Motion}}

\rev{
The current experiments use a 6-DOF Stewart platform to reproduce predefined base motions and therefore cannot capture the effects of manipulator reaction forces and moments on the base. 
This approximation is reasonable when the mass and inertia of the ship are much greater than those of the manipulator and payload.
However, the motion of a heavy manipulator may affect the attitude of a small ship. Future work may incorporate the coupled dynamics and use the manipulator redundancy to reduce the reaction moments without compromising the end-effector task.
}

\section{Conclusion}
This paper \rev{presents} an optimization-based inverse dynamics control framework for ship-borne manipulators operating under base motion. By explicitly compensating for the inertial effects induced by the floating base, the method enables accurate trajectory tracking and compliant interaction with the environment. To support reliable feedforward compensation, an ESKF is developed to estimate the base motion by fusing IMU measurements with end-effector pose feedback.

Both simulation and real-world experiments demonstrate the effectiveness of the framework. The results show a significant improvement in trajectory tracking accuracy compared with baseline approaches. Furthermore, the inherent task-space compliance enables reliable execution of a dynamic peg-in-hole insertion task, reducing contact forces and improving task success rates.

Future work will investigate integrating adaptive control techniques to further mitigate modeling errors. In addition, incorporating onboard perception and multi-sensor fusion will be explored to enable fully autonomous manipulation in dynamic maritime environments.

%% file: appendix.tex
\appendix
\subsection{Uniform Ultimate Boundedness Analysis}
\label{app:uub_proof}

{\color{R1}The constrained task-space error dynamics in
\eqref{eq:constrained_impedance_error} can be written as
\begin{equation}
    \ddot{\boldsymbol{e}}
    +
    \boldsymbol{K}_d\dot{\boldsymbol{e}}
    +
    \boldsymbol{K}_p\boldsymbol{e}
    =
    \boldsymbol{w},
    \label{eq:appendix_error_input}
\end{equation}
where
\begin{equation}
    \boldsymbol{w}
    =
    -
    \boldsymbol{\Lambda}^{-1}\boldsymbol{F}_{\mathrm{ext}}
    -
    \boldsymbol{J}\boldsymbol{M}^{-1}\boldsymbol{\tau}_d
    -
    \boldsymbol{J}\Delta\ddot{\boldsymbol{q}}_c .
    \label{eq:appendix_w}
\end{equation}

Assume that the joint positions and velocities remain within
their finite limits, the manipulator remains away from
kinematic singularities, and the QP remains feasible. The
reference trajectory, base velocity and acceleration, external
wrench, and unmodeled disturbance are assumed to be bounded.
Under these conditions, the kinematic and dynamic quantities
appearing in the controller are bounded. In particular,
$\ddot{\boldsymbol{q}}_{\mathrm{ns}}$ and the unconstrained
solution $\ddot{\boldsymbol{q}}_0$ are bounded. Since the QP
solution $\ddot{\boldsymbol{q}}^{*}$ satisfies finite
joint-acceleration limits,
\begin{equation}
    \Delta\ddot{\boldsymbol{q}}_c
    =
    \ddot{\boldsymbol{q}}^{*}
    -
    \ddot{\boldsymbol{q}}_0
\end{equation}
is also bounded. Therefore, there exists a positive constant
$\bar{w}$ such that
\begin{equation}
    \left\|\boldsymbol{w}\right\|
    \leq
    \bar{w}.
    \label{eq:appendix_w_bound}
\end{equation}

Define
\begin{equation}
    \boldsymbol{z}
    =
    \begin{bmatrix}
        \boldsymbol{e}^{T} &
        \dot{\boldsymbol{e}}^{T}
    \end{bmatrix}^{T}.
\end{equation}
Equation \eqref{eq:appendix_error_input} becomes
\begin{equation}
    \dot{\boldsymbol{z}}
    =
    \boldsymbol{A}_e\boldsymbol{z}
    +
    \boldsymbol{B}_e\boldsymbol{w},
\end{equation}
where
\begin{equation}
    \boldsymbol{A}_e
    =
    \begin{bmatrix}
        \boldsymbol{0} & \boldsymbol{I} \\
        -\boldsymbol{K}_p & -\boldsymbol{K}_d
    \end{bmatrix},
    \qquad
    \boldsymbol{B}_e
    =
    \begin{bmatrix}
        \boldsymbol{0} \\
        \boldsymbol{I}
    \end{bmatrix}.
\end{equation}
Since $\boldsymbol{K}_p$ and $\boldsymbol{K}_d$ are positive
definite, $\boldsymbol{A}_e$ is Hurwitz. For any positive
definite matrix $\boldsymbol{Q}_L$, there exists a positive
definite matrix $\boldsymbol{P}$ satisfying
\begin{equation}
    \boldsymbol{A}_e^{T}\boldsymbol{P}
    +
    \boldsymbol{P}\boldsymbol{A}_e
    =
    -
    \boldsymbol{Q}_L.
\end{equation}

Consider the Lyapunov function
\begin{equation}
    V
    =
    \boldsymbol{z}^{T}\boldsymbol{P}\boldsymbol{z}.
\end{equation}
Its derivative satisfies
\begin{equation}
\begin{aligned}
    \dot{V}
    &=
    -
    \boldsymbol{z}^{T}\boldsymbol{Q}_L\boldsymbol{z}
    +
    2\boldsymbol{z}^{T}
    \boldsymbol{P}\boldsymbol{B}_e\boldsymbol{w} \\
    &\leq
    -
    \lambda_{\min}(\boldsymbol{Q}_L)
    \left\|\boldsymbol{z}\right\|^2
    +
    2
    \left\|\boldsymbol{P}\boldsymbol{B}_e\right\|
    \left\|\boldsymbol{z}\right\|
    \bar{w}.
\end{aligned}
\end{equation}
Thus, for any $\theta\in(0,1)$, $\dot{V}<0$ whenever
\begin{equation}
    \left\|\boldsymbol{z}\right\|
    >
    \frac{
        2\left\|\boldsymbol{P}\boldsymbol{B}_e\right\|
    }{
        \theta\lambda_{\min}(\boldsymbol{Q}_L)
    }
    \bar{w}.
\end{equation}
Therefore, the error state is uniformly ultimately bounded, with the conservative bound
\begin{equation}
    \limsup_{t\rightarrow\infty}
    \left\|\boldsymbol{z}(t)\right\|
    \leq
    \sqrt{
        \frac{
            \lambda_{\max}(\boldsymbol{P})
        }{
            \lambda_{\min}(\boldsymbol{P})
        }
    }
    \frac{
        2\left\|\boldsymbol{P}\boldsymbol{B}_e\right\|
    }{
        \theta\lambda_{\min}(\boldsymbol{Q}_L)
    }
    \bar{w}.
\end{equation}
Hence, both $\boldsymbol{e}$ and $\dot{\boldsymbol{e}}$ are uniformly ultimately bounded.
}